\documentclass[10pt, a4paper]{dukenlp}

\usepackage{hyperref}
\usepackage{url}

\usepackage{natbib}
\usepackage{times}
\usepackage{subcaption}
\usepackage{latexsym}
\usepackage{graphicx}
\usepackage{amsmath}
\usepackage{amsfonts}
\usepackage{bbm}
\usepackage{listings}
\usepackage{booktabs}
\usepackage{multirow}
\usepackage{comment}
\usepackage{xspace}
\usepackage{dialogue}
\usepackage{algorithm}
\usepackage{algorithmic}
\usepackage{array}
\usepackage{tabularx}

\usepackage{wrapfig}      
\usepackage{float}        
\usepackage[most]{tcolorbox}
\usepackage{caption}      

\definecolor{darkblue}{RGB}{0,51,102}

\tcbset{
  promptbox/.style={
    enhanced,
    colback=gray!3,
    colframe=darkblue,
    boxrule=0.5pt,
    left=6pt, right=6pt, top=6pt, bottom=6pt,
    before upper=\setlength{\parskip}{3pt}\setlength{\parindent}{0pt},
    sharp corners,
    breakable=false,
    before skip=0pt, after skip=0pt,
  }
}

\setcitestyle{authoryear,round,citesep={;},aysep={,},yysep={;}}

\usepackage{amsmath,amsfonts,bm}

\def\eqref#1{equation~\ref{#1}}

\def\1{\bm{1}}

\DeclareMathAlphabet{\mathsfit}{\encodingdefault}{\sfdefault}{m}{sl}
\SetMathAlphabet{\mathsfit}{bold}{\encodingdefault}{\sfdefault}{bx}{n}

\usepackage{xurl}
\usepackage{capt-of}
\usepackage{placeins}
\usepackage{tikz}
\usepackage[normalem]{ulem}

\usetikzlibrary{arrows.meta,positioning,fit,calc}

\title{Multimodal AI Detection for Math-Heavy\\Scientific Text}


\author{
Yixuan Huang$^{1}$, Danish Pruthi$^{2}$, Bhuwan Dhingra$^{1}$ \\
$^{1}$Duke University, $^{2}$Indian Institute of Science\\
\texttt{yixuan.huang@duke.edu, bdhingra@cs.duke.edu}
}

\fancypagestyle{firstpagenote}[firststyle]{%
  \fancyfoot[L]{\footerfont\hypertarget{Hfootnote.1}{}\textsuperscript{1}The Pangram~4 report explicitly states that mathematical prompts are excluded from its evaluation.}%
}

\begin{document}

\maketitle
\suppressfloats[t]   

\begin{abstract}

As large language models become increasingly capable of
generating, and assisting with, scientific writing,
there is a need for greater transparency and disclosure around the use
of AI in research.
Consequently, 
a veritable  
subfield of research and a number of commercial services 
aim to flag AI-generated content automatically.
However, existing detectors which rely only on text
extracted from PDFs fail to benefit
from the rich visual information contained in the formatting of
\emph{math-heavy} scientific content.
To study this, we select human-written proof and prose passages from pre-2022 arXiv LaTeX sources and re-generate them using $15$ modern LLMs within the context of the same paper.
This leaves us with coupled human and AI outputs in three different representations: the LaTeX source, a rendered image of the PDF, and the text extracted from the PDF.
We find that the accuracy of the state-of-the-art Pangram 4
drops significantly as the mathematical
content in a passage increases,
especially when evaluated on text as opposed to LaTeX.
However, this is not an inherent limitation of detecting AI-generated math:
finetuning Qwen-3.5-27B based detectors on this domain recovers the gap,
with LaTeX inputs performing significantly better than text since
they preserve more formatting information.
Since LaTeX sources are rarely available in practice,
we also train \textsc{SnapJudge}:
an image-based detector operating on screenshots of both prose
and proofs from papers. 
At $0.1\%$ false positive rate, \textsc{SnapJudge}
achieves a detection rate of $92\%$ on proofs and $93\%$ on prose generated by frontier models, substantially outperforming
both Pangram 4 and our text-based detector.
We release our data, models and code at \url{https://nlplab.cs.duke.edu/snapjudge}.

\end{abstract}
\thispagestyle{firstpagenote}

\begin{figure}[t]
\vspace{9pt}
\centering
\resizebox{0.65\linewidth}{!}{%
\begin{tikzpicture}[
  panel/.style={draw=black!30, inner sep=2pt, rounded corners=1pt},
  ours/.style={draw=blue!60!black, semithick, inner sep=2pt, rounded corners=1pt},
  ttl/.style={font=\small\bfseries, inner sep=1pt},
  meth/.style={font=\normalsize\bfseries, draw, rounded corners=3pt, inner sep=4pt, align=center},
  arr/.style={-{Stealth[length=2.2mm]}, semithick, black!55, rounded corners=2pt},
  marr/.style={-{Stealth[length=2.2mm]}, semithick, rounded corners=2pt},
  albl/.style={font=\small\itshape, fill=white, inner sep=1.5pt}
]
\node[panel, anchor=north] (src) at (0,0.09)    {\includegraphics[width=3.600cm, trim=4.0 8.7 13.1 8.7, clip]{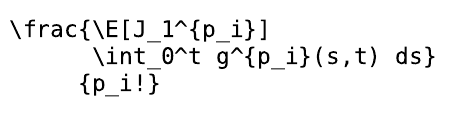}};
\node[panel, anchor=north] (txt) at (3.2,0.09) {\includegraphics[width=1.561cm, trim=1.75 10.0 15.7 10.0, clip]{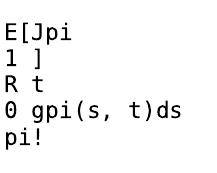}};
\node[ours,  anchor=north] (img) at (5.75,0.09) {\includegraphics[width=2.427cm, trim=3.4 3.9 3.1 3.8, clip]{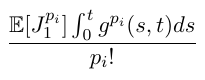}};
\node[ttl] (tsrc) at (0,0.33)    {LaTeX};
\node[ttl] (ttxt) at (3.2,0.33) {Text};
\node[ttl] (timg) at (5.75,0.33) {Image};
\node[draw=black!55, rounded corners=2pt, font=\small\bfseries,
      minimum width=1cm, minimum height=0.5cm] (pdf) at (3.2,1.05) {PDF};
\draw[arr] (tsrc.north) to[bend left=12] node[albl, pos=0.45] {compile} (pdf.west);
\draw[arr] (pdf.south) -- node[albl, pos=0.5, anchor=west, xshift=0.6mm, yshift=-1mm] {extract} ([yshift=0.5mm]ttxt.north);
\draw[arr] (pdf.east) to[bend left=12] node[albl, pos=0.55] {render} ([yshift=0.5mm]timg.north);
\coordinate (row) at (0,-2.06);
\node[meth, color=black!70, anchor=west, align=center] (ex) at (src.west |- row) {Existing\\[-1pt]detectors};
\tikzset{chip/.style={font=\small\bfseries, align=center, draw=blue!60!black, rounded corners=2pt, inner sep=2.5pt, text=blue!60!black}}
\node[chip, fill=blue!12, anchor=east] (cS) at ([xshift=-3.6pt]img.east |- row) {SnapJudge\\[-1pt]\textnormal{\footnotesize(image)}};
\node[chip] (cT) at ([xshift=-0.3cm]txt.east |- row) {Text-based\\[-1pt]detector};
\node[chip] (cL) at ([xshift=-0.45cm]src.east |- row) {LaTeX\\[-1pt]detector};
\node[draw=blue!60!black, semithick, rounded corners=3pt, fit=(cL)(cT)(cS), inner sep=3pt] (ms) {};
\node[font=\small\bfseries, text=blue!60!black, anchor=north] at (ms.south) {Our detectors};
\draw[arr, dashed] (ex.north |- src.south) -- (ex.north);
\draw[arr] ([yshift=0.025cm]txt.south west) -| ([xshift=0.55cm]ex.north);
\draw[marr, blue!60!black, dashed, preaction={draw=white, line width=3pt, solid, -}] (cL.north |- src.south) -- (cL.north);
\draw[marr, blue!60!black] (cT.north |- txt.south) -- (cT.north);
\draw[marr, blue!60!black, thick] (cS.north |- img.south) -- (cS.north);
\node[albl, text=red!70!black] at ($(ex.north |- src.south)!0.5!(cL.north |- src.south)+(0,-0.215)$) {$\times$~usually unobtainable};
\end{tikzpicture}}
\vspace{-6pt}
\caption{\textbf{The same mathematical expression shown as LaTeX source, PDF-extracted text, and a rendered image.}
We study AI detection across three representations and find that rendered images generally yield higher detection rates than extracted text and approach the performance obtained with the original LaTeX source.}
\label{fig:textdamage}
\vspace{-10pt}
\end{figure}

\section{Introduction}

Mathematical capabilities of LLMs have risen from solving middle-school math problems \citep{cobbe2021training} to winning math olympiads \citep{deepmind2025imo}, and now tackling longstanding open problems,
such as the Jacobian conjecture \citep{tao2026jacobian}
and (allegedly) the Navier--Stokes Millennium Prize problem \citep{openai2026navierstokes}.
These advances highlight the urgent need for improving transparency about the use of AI
when publishing scientific papers,
including the technical content such as proofs
and derivations that support their claims.
Even when the claims in a paper are correct, understanding how they were produced still matters for evaluating researchers' contributions and assigning credit.
This need is reflected in publication policies: many conferences and publishers now require authors to disclose their use of AI \citep{neurips2025llm,iclr2026llm,neurips2026aigen,springernature2026ai, mohapatra2026expectations}.
Such policies rest on honest disclosure, which can only be verified, and therefore enforced, when accurate detection is available.
Although the state-of-the-art commercial detector, Pangram 4 \citep{pangram4report},  
is highly accurate at flagging  general-purpose AI content, we find that
its effectiveness degrades significantly on
scientific text, especially math-heavy text,
which comes with rich visual formatting.\footnotemark

For ordinary prose, we can expect that any given idea can be expressed in many different ways, and the choices an LLM makes among them leave traces that can be detected \citep{detectgpt,binoculars,liang2025nhb}.
However, it is unclear whether such detectable signs remain in the much more constrained and formal setting of math-heavy scientific writing. 
In this work, we present the first systematic study of AI text detection on math-heavy scientific writing under $3$ different representations: LaTeX, text and rendered images of PDF pages (Figure~\ref{fig:textdamage}).
We find that \textbf{AI-generated math remains highly detectable under all three representations}.

LaTeX is the dominant language for typesetting math in scientific papers \citep{arxiv2026html},
however, in practice readers seldom have access to the underlying LaTeX
source of a paper they are reading.
Prior detectors deal with PDFs by either:
(i) extracting text from the PDF, using tools such as PyMuPDF \citep{pymupdf},
which severely damages formatting of the mathematical content and, as we show, reduces detection accuracy
(Section~\ref{sec:Results});
or (ii) running OCR on the page images using specialized tools
such as Marker and Tesseract \citep{marker, smith2007overview},
which preserve math but introduce additional formatting signals that impede detection 
\citep{pangram2025iclr}.

Hence, we develop \textsc{SnapJudge}, an image-based detector that identifies AI-generated writing directly from rendered scientific passages.
This requires minimal preprocessing and preserves the stylistic choices behind math writing
which can aid detection.
Specifically,
we construct our data
by
extracting the proof and prose passages from LaTeX sources of pre-2022 arXiv papers,
and re-generating them using modern LLMs (conditioned on the rest of the paper).
Using the resulting data, we finetune three Qwen3.5-based detectors \citep{qwen35}, one per representation,
and show that while the LaTeX detector performs best, the image-based detector has comparable
accuracy and also
outperforms the text-based detector (Section~\ref{sec:Results}).
On the same test data, we find that Pangram~4 only detects $36\%$ of the AI-generated proofs
when tested on the LaTeX source,
with even worse performance upon using the extracted text.
Overall, our detectors lead to a ${>}90\%$ detection rate at a relatively strict threshold
of $0.1\%$ false positive rate,
and also
hold up outside mathematics.
Despite concerns that supervised detectors collapse on unfamiliar domains \citep{rethinking2026}, 
\textsc{SnapJudge} performs excellently on MIRAGE \citep{mirage}, RAID \citep{raid}, AI-written peer reviews \citep{intellabsbench}, and the EditLens dataset \citep{editlens} (Section~\ref{sec:ood}).

When we reduce the false-positive rate to $0.01\%$, \textsc{SnapJudge} still detects more than $82\%$ of the AI-written passages (Figure~\ref{fig:fpr-tpr}), highlighting the viability of our tool for practical use.
We further analyze what makes AI-generated math detectable,
and invalidate explanations 
that conjecture
that AI-generated math is qualitatively different or may contain more errors, and therefore is detectable (Section~\ref{sec:mathfrac}).
We also investigate if prose around the mathematical formulas gives away signs of AI writing,
and find
that passages do become harder to detect as 
the percentage of 
formulas in them increases,
but we are still able to detect over $80\%$ proofs which are
composed primarily of formulas and equations.


To summarize, our contributions are the following:

\begin{enumerate}
\item We contribute a dataset containing math-heavy text, consisting of 47K human-AI paired proofs and 45K paired prose passages, where each pair includes three representations, namely the LaTeX source, the text extracted from the PDF, and the rendered image.
\item We show that existing zero-shot and open-source detectors fail to detect proofs from frontier models, and even the state-of-the-art Pangram~4 detects only 32.4\% of them (from extracted text), or 65.6\% upon recalibration.

\item We systematically study the detection performance under three different representations, and find that detectors operating on the LaTeX source perform best, whereas text-based detectors struggle with math-heavy text. Image-based detectors are comparable to the performance of detectors that use LaTeX source. 
\item We propose \textsc{SnapJudge}, an image-based text detector, which also supports estimating the proportion of the text in an image that was AI-generated. 
It is the most effective practical detector for math-heavy text, outperforming the text-based Pangram~4 by $26.2$ percentage points on frontier-model proofs.
\item We analyze AI usage trends in arXiv papers across fields and find that AI entered mathematics relatively late, but its use in mathematics is now growing faster and faster, with the share of mathematics passages detected as AI-generated rising from essentially zero before 2023 to $23.1\%$ of prose and $17.6\%$ of proofs in August 2026.
Proofs still lag behind prose but are catching up quickly.

\end{enumerate}


\section{Related Work}
\label{gen_inst}

\paragraph{AI-Generated Text Detection.}
Post-hoc detectors of AI-generated text either score a text using statistics from a a language model, such as its likelihood or perplexity \citep{detectgpt,fastdetectgpt,binoculars}, or train a classifier using labeled human-written and AI-generated examples \citep{editlens,pangram4report}.
The commercial Pangram~4 is the strongest and most widely used detector of the second kind, with a detection rate above $99\%$ at a false-positive rate of $0.004\%$ on its evaluation \citep{pangram4report}. Pangram, along with EditLens, go beyond a binary label by locating AI-written words or estimating the extent of AI editing \citep{pangram4report,editlens}.
These detectors take text as input and are primarily developed and evaluated on natural-language prose.
In math-heavy scientific writing, however, much of the information lies in the formatting of formulas, which may pose a challenge to such detectors.
We evaluate the off-the-shelf detectors and find their performance to be limited, most markedly on proofs (Section~\ref{sec:main}).

\paragraph{Documents as Images.}
Our work builds on the recent trend of using 
vision-language models to process rendered images of text, a capability which has been steadily improving \citep{qwen25vl,ocrbench}.
PIXEL showed that language modeling can be built directly on the pixels of rendered text \citep{pixel}, ColPali retrieves documents from their page images without any text extraction \citep{colpali}, and Glyph and LensVLM render long text into images to compress the context of a model \citep{glyph,lensvlm}.
For scientific document retrieval, \citet{docimage_fallshort} find that whole-page image representations underperform the LaTeX-source form of the text and degrade more sharply as documents grow longer, which is consistent with our conclusion that the LaTeX source is the easiest representation to detect.

\paragraph{AI Detection from Images.}
Existing image-based detectors target whether the image itself was synthesized using AI, rather than whether the text shown in an image was written by AI \citep{wang2020cnn,ntire2026,pangramimage2026,textfake2026,textrich2026}.
To our knowledge, the only other approach studying the latter setup is the concurrent work AIDoc \citep{aidoc2026}.
However, its evaluation is limited to earlier open-weight generators, with reported AUROC scores of only $0.81$-$0.87$, and does not establish effective detection at the low false-positive rates needed for practical deployment.
Our work focuses on math-heavy scientific content, comparisons across input modalities, and detecting text generated by frontier models at extremely low false-positive rates.
Another existing practice involves using OCR to recover text from images and then running detection on the result.
Our experiments show that such OCR-recovered text embeds its own signatures and is difficult to detect for math-heavy content (Section~\ref{sec:main}).

\section{Setup \& Methods}
\label{headings}

\subsection{Dataset}\label{sec:dataset}

\noindent \textbf{Source material.} We extract texts from the LaTeX source of arXiv papers submitted between 2019 and 2021 (before ChatGPT was released) and never revised afterwards.
This pool serves as our ``Human'' reference set and contains two kinds of texts: proofs and prose.
A proof passage is the body of the proof of a theorem, lemma, proposition or corollary, taken from the LaTeX source as the text between \verb|\begin{proof}| and \verb|\end{proof}|. Prose is a paragraph that may contain inline mathematics and other LaTeX content such as references, but no theorem, proof, figure or table. 

\noindent \textbf{Paired regeneration.} We flatten all LaTeX files of each paper into a single file that contains the complete LaTeX source. 
For each proof or prose passage, we prompt an LLM with the entire source of the paper except the target passage, and ask it to produce the LaTeX source of the missing passage in the style of the original paper.
Each passage is regenerated by one of 15 LLMs: ten open-source models run locally with vLLM \citep{vllm} and five frontier models accessed through APIs (Appendix~\ref{app:prompts}).
This results in a paired dataset of passages, where each passage has three modalities or representations:
(1) the LaTeX source;
(2) the PDF-extracted text, where each passage is compiled into its own PDF with the preamble and the macro definitions coming from the source paper, and the text is then extracted with PyMuPDF;
and (3) the rendered image of the same PDF.
We discard $2.4\%$ of AI-generated proofs that are shorter than $50$ words.
Compiling each passage on its own with the preamble of its source paper succeeds for $94\%$ of the pairs, which constitute the text and image versions of the dataset.


\noindent \textbf{Real pages.} We construct a variant that, we believe, is closer to real-world use.
For each target passage, we place the model-generated content back at its original position in the flattened source and recompile the whole paper.
This yields a real page on which the AI-generated passage is surrounded by human-written ones (an example is shown in Figure~\ref{fig:fraction-page} in Appendix~\ref{app:fraction-confusion}).
On these pages we crop windows that are as wide as the column and fully cover the target passage together with portions of human-written context.
The AI fraction $p$ of a window is defined as the ratio of the height of the target-passage text to the total height of the text within the window.
A crop that covers exactly the target passage has $p=1$ on the page with the AI-generated version and $p=0$ on the page with the human-written original, and for the remaining windows $p$ is continuously distributed in $(0,1)$.
We use this data to further train a \emph{fraction estimator}, i.e., a model that predicts how much text in a window is AI (Appendix~\ref{app:fraction-confusion}).
Those that contain only AI-generated text ($p=1$) or only human-written text ($p=0$) are also added to the training data of the text-based detector and \textsc{SnapJudge}, so that the detectors become more robust to real-page artifacts such as line numbers and watermarks.

\noindent \textbf{Multi-level AI-editing test set.} 
To evaluate detection under varying degrees of AI editing, 
we construct a multi-level AI-editing test set from 1,500 human-written proofs and prose passages each, divided equally among the ten open-source generators.
Each passage undergoes three editing exercises: sentence-level editing, whole-passage rewriting, and regeneration from an outline of its logical steps.
Prompts are provided in Appendix~\ref{app:prompts}.

\begin{wraptable}{R}{0.41\textwidth}
\vspace{-\baselineskip}
\centering
\footnotesize
\setlength{\tabcolsep}{2.5pt}
\begin{tabular}{@{}lcccc@{}}
\toprule
 & Train & Val & Test & Calib. \\
\midrule
LaTeX & 34.1k/30.2k & 1.3k/2.2k & 11.5k/12.2k & 357.4k \\
Text & 32.3k/30.1k & 1.1k/1.9k & 9.9k/11.9k & 354.0k \\
Image & 31.6k/30.0k & 1.1k/1.7k & 9.9k/11.8k & 351.6k \\
\bottomrule
\end{tabular}
\vspace{-4pt}
\caption{Dataset size: AI--human pairs (proofs\,/\,prose) per split, and human-written passages in the calibration pools.}
\label{tab:data}
\vspace{-9pt}
\end{wraptable}

\noindent \textbf{Splits and calibration pool.} 
We split training, validation and test data by paper, and Table~\ref{tab:data} lists their sizes.
All validation pairs and 98.6\% of the training pairs come from papers of 2019 and 2020, while all test pairs and the remaining 1.4\% of the training pairs come from papers of 2021.
For most passages, the training sets of the text-based detector and \textsc{SnapJudge} additionally include the real-page versions described above.
To set thresholds at low false-positive rates, we build calibration pools of about 350K human-written passages from papers of 2018 to 2020 (Table~\ref{tab:data}), one pool per representation. Each pool consists mostly of passages drawn from the body of papers, avoiding references, figures, tables, headers and footers, together with around 30K proofs.


\noindent \textbf{Note on contamination.} We acknowledge that source papers are likely included in LLM pretraining corpora, creating a risk of data contamination during both generation and subsequent detection \citep{magar-schwartz-2022-data}. Specifically, there are two concerns: first, it is possible that 
AI models generate proofs that are verbatim reproductions of human-written ones, or are highly similar to them.
To avoid such concerns during generation, we remove the $0.3\%$ of AI-generated passages whose LaTeX source has more than $10\%$ word-level 8-gram overlap with that of the original passage.
Second, it is possible that 
any detector finetuned 
to flag AI-generated content 
may partly rely on whether its backbone encountered the text during pretraining.
Unfortunately, this contamination risk is shared by all supervised detectors (including the baselines we compare against). 
Further, unlike other applications, contamination concerns are harder to alleviate for AI text detection, 
as there is simply no guarantee that any online text produced since 2022 is completely human written, so 
most detectors rely on pre-2022 content.

\subsection{Detectors}\label{sec:detectors}

We train a separate detector for each of the three representations, referred to as the LaTeX detector, the text-based detector and the image-based detector, which we call \textsc{SnapJudge}, as well as an AI-fraction estimator that complements \textsc{SnapJudge} on page images by predicting how much of an image is AI-generated (Appendix~\ref{app:fraction-confusion}).
Each detector treats detection as binary classification at the passage level. Given one passage, it outputs the probability that the passage was generated by an LLM, and the passage is flagged as AI-generated when this score exceeds a threshold.

\noindent \textbf{Backbone and training.} The detectors use Qwen3.5 \citep{qwen35} as the backbone.
We fine-tune the attention and feed-forward projection matrices with LoRA \citep{lora} (the hyperparameters are given in Appendix~\ref{app:hparams}), and attach a linear classification head to the hidden state of the last token of the sequence. Training is done with the cross-entropy loss.
The LaTeX detector and the text-based detector cut the input into overlapping windows of $512$ tokens with a stride of $256$ tokens, each window is one training example and inherits the label of the whole passage.
\textsc{SnapJudge} takes the whole rendered image as input.
An image is encoded into visual tokens that each cover a $32{\times}32$ pixel patch, with a budget of about $1.25$ megapixels in total. 
Around $6\%$ of the images exceed this budget and are down-scaled proportionally before being fed to the model.
Each detector is trained twice with different random seeds but otherwise identical settings.
Reported detection rates are averaged across the two independently calibrated runs.

\noindent \textbf{Scoring.} At detection time, the LaTeX detector and the text-based detector divide the input into windows in the same way as in training.
The passage score is the mean AI-class probability over its text windows or page images.
All detection rates are obtained by thresholding the scores of the detector at a common false-positive rate on the human calibration pool described in Section~\ref{sec:dataset}.
We assess score calibration in Appendix~\ref{app:reliability}.

\subsection{Baselines}\label{sec:baselines}
For comparison, we select several off-the-shelf detectors and train our own shallow baselines.
The shallow baselines are logistic-regression classifiers trained on the same examples as our other detectors, separately for each representation and for proofs and prose.
We train three of them, using stylometric features (including sentence length, the frequency of punctuation and symbols, and vocabulary richness), TF-IDF features of character 3- to 5-grams, and TF-IDF features of word unigrams and bigrams, respectively.
Among the off-the-shelf detectors, we select Binoculars \citep{binoculars} and Fast-DetectGPT \citep{fastdetectgpt} to represent zero-shot methods, which score a text by the difference in its likelihood under different language models.
As supervised detectors, we choose the open-source EditLens \citep{editlens} and the state-of-the-art Pangram~4 \citep{pangram4report}.
Pangram is regarded as the most accurate detector available and is widely used to measure AI use in the wild, for example in newspaper articles \citep{russell2025newspapers}, arXiv papers \citep{elazar2026arxiv} and self-published books \citep{chakrabarty2026books}.

\section{Results}\label{sec:Results}

\subsection{Detection rates under the three representations}
\label{sec:main}
In Table~\ref{tab:unified-main}, we report the detection rates at a low false positive rate of $0.1\%$ computed on a test set.
For every detector, except Pangram, we set the threshold using the calibration pool of Section~\ref{sec:dataset}.
For Pangram~4 we report both its own returned labels (AI and mixed are both counted as AI) and the results obtained by re-thresholding its returned AI assistance score at a $0.1\%$ false-positive rate.
Owing to the Pangram API budget, we score only 2,500 proof pairs and 500 prose pairs, and its threshold for the $0.1\%$ false-positive rate is determined on these 3,000 human-written passages.
\begin{table}[t]
\centering
\footnotesize
\renewcommand{\arraystretch}{1.0}
\setlength{\tabcolsep}{0pt}
\begin{tabular*}{\linewidth}{@{\extracolsep{\fill}}lcccc@{\hskip 14pt}c@{\hskip 0pt}cccc@{}}
\toprule
 & \multicolumn{4}{c}{\textbf{LaTeX}} & & \multicolumn{4}{c}{\textbf{Text} } \\
\cmidrule(lr){2-5}\cmidrule(lr){7-10}
 & \multicolumn{2}{c}{Open avg.} & \multicolumn{2}{c}{Frontier avg.} & & \multicolumn{2}{c}{Open avg.} & \multicolumn{2}{c}{Frontier avg.} \\
\cmidrule(lr){2-3}\cmidrule(lr){4-5}\cmidrule(lr){7-8}\cmidrule(lr){9-10}
 & Prose & Proofs & Prose & Proofs & & Prose & Proofs & Prose & Proofs \\
\midrule
\multicolumn{10}{c}{\emph{Shallow baselines (features + logistic regression)}} \\
\midrule
stylometric & 0.030 & 0.003 & 0.014 & 0.000 & & 0.014 & 0.002 & 0.006 & 0.000 \\
char TF-IDF & 0.558 & 0.394 & 0.288 & 0.100 & & 0.480 & 0.317 & 0.238 & 0.145 \\
word TF-IDF & \underline{0.664} & \underline{0.584} & \underline{0.359} & \underline{0.308} & & \underline{0.559} & \underline{0.485} & \underline{0.247} & \underline{0.238} \\
\midrule
\multicolumn{10}{c}{\emph{Off-the-shelf detectors}} \\
\midrule
Binoculars & 0.101 & 0.093 & 0.002 & 0.006 & & 0.002 & \underline{0.037} & 0.001 & 0.001 \\
Fast-DetectGPT & 0.154 & \underline{0.122} & 0.011 & 0.013 & & 0.094 & 0.008 & 0.006 & 0.001 \\
EditLens-Llama & \underline{0.657} & 0.086 & 0.340 & 0.007 & & \underline{0.525} & 0.014 & 0.245 & 0.000 \\
EditLens-RoBERTa & 0.268 & 0.093 & 0.116 & 0.022 & & 0.191 & 0.021 & 0.082 & 0.008 \\
\addlinespace[2.5pt]
Pangram v3.3.2 & -- & -- & 0.581 & 0.134 & & -- & -- & -- & -- \\
\quad\textit{recalibrated} & -- & -- & 0.742 & 0.368 & & -- & -- & -- & -- \\
\addlinespace[2.5pt]
Pangram 4 & -- & -- & 0.797 & 0.363 & & -- & -- & 0.694 & 0.324 \\
\quad\textit{recalibrated} & -- & -- & \underline{0.908} & \underline{0.875} & & -- & -- & \underline{0.871} & \underline{0.656} \\
\quad\textit{OCR} & -- & -- & -- & -- & & -- & -- & 0.660 & 0.250 \\
\quad\textit{OCR, recalibrated} & -- & -- & -- & -- & & -- & -- & 0.860 & 0.500 \\
\midrule
\multicolumn{10}{c}{\emph{Ours, 27B}} \\
\midrule
LaTeX / text-based & \textbf{\underline{0.943}} & \textbf{\underline{0.920}} & \textbf{\underline{0.957}} & \textbf{\underline{0.938}} & & 0.921 & 0.889 & 0.920 & 0.877 \\
\quad\textit{OCR (Tesseract)} & -- & -- & -- & -- & & 0.892 & 0.867 & 0.885 & 0.861 \\
\textsc{SnapJudge} & -- & -- & -- & -- & & \textbf{\underline{0.932}} & \textbf{\underline{0.933}} & \textbf{\underline{0.928}} & \textbf{\underline{0.918}} \\
\bottomrule
\end{tabular*}

\caption{Detection rate at $0.1\%$ FPR on the LaTeX source (left) and on inputs derived from the PDF (right), i.e, extracted text,  OCR text (\emph{OCR}) or the rendered page (\textsc{SnapJudge}). \emph{Open avg.} and \emph{Frontier avg.} average over the ten open-source and the five frontier models. When the LaTeX source is not available, \textsc{SnapJudge} is the most accurate detector, especially on frontier proofs. Owing to API cost, we evaluate Pangram on frontier models and its v3.3.2 only on LaTeX source.}
\vspace{-5pt}

\label{tab:unified-main}

\end{table}

The shallow baselines perform poorly, especially on proofs, where they still rely mostly on the text of a proof rather than mathematical symbols.
Removing all mathematical symbols (inline and otherwise) from the texts and retraining the shallow baselines lowers their detection rate by only about $2\%$ on proofs and $1\%$ on prose, whereas keeping only these mathematical symbols makes them fail almost completely, suggesting that these baselines do not rely on the mathematical content. 
These shallow classifiers are not viable for detection, as the best of them detect only $0.308$ of frontier proofs at a $0.1\%$ false-positive rate.

Among the off-the-shelf detectors, only Pangram~4 clearly outperforms the shallow baselines.
The two zero-shot detectors, Binoculars and Fast-DetectGPT, fail almost completely.
EditLens detects part of the prose ($0.245$ on the text) but fails to detect proofs accurately.
The state of the art, Pangram~4, detects only $0.324$ of the proofs when given the text extracted from the PDF according to its own labels returned from the API, and even classifies $1$ human-written passage as AI-assisted.
After threshold recalibration, it's performance is $0.656$, far below the ${>}0.99$ accuracy that it reports in other domains \citep{pangram4report}.

Of our own detectors, the LaTeX detector performs best, \textsc{SnapJudge} comes close to it, and the text-based detector performs worst.
This is in line with our expectation that LaTeX preserves the most formatting and text the least.
We also experiment with an OCR-based approach, where we take the image representations of the passages and extract the text with two OCR tools, Tesseract \citep{smith2007overview} and Marker \citep{marker}.
Neither tool works well, because OCR can add symbols that AI models commonly use to human-written text \citep{pangram2025iclr}, which raises the detection threshold and thus lowers the detection rate.
Table~\ref{tab:unified-main} reports the results of Tesseract, which performs somewhat better.
On the representations available in practice, namely the text and the image, the frontier-family average of \textsc{SnapJudge} ($0.918$ and $0.928$) is far above the $0.656$ and $0.871$ that Pangram~4 reaches after calibration. Pangram~4 is especially weak on short passages, whereas our detectors remain accurate even below 100 words (Appendix~\ref{app:length}).

\begin{figure}[t]
\centering
\includegraphics[width=\textwidth]{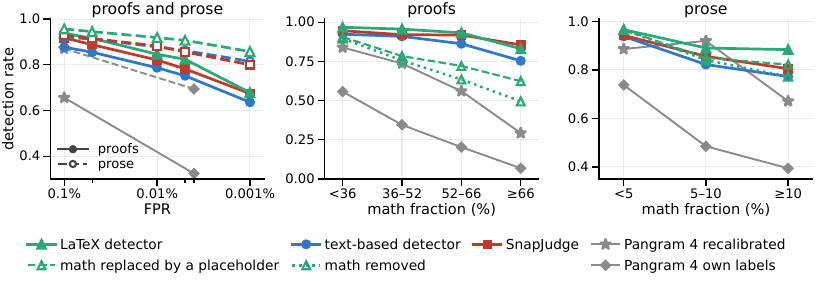}
\vspace{-20pt}
\caption{Detection rate of our 27B detectors on frontier passages. The left panel tightens the FPR from $0.1\%$ to $0.001\%$ for proofs (solid) and prose (dashed), and Table~\ref{tab:fpr-tiers} lists the values.
Pangram~4 is shown after recalibration at $0.1\%$ and with its own labels at $0.004\%$, the FPR reported in its technical report. The right two panels split the passages by math fraction at $0.1\%$ FPR, where the dashed and dotted green lines show the LaTeX detector on the same passages with every math segment replaced by a placeholder or with the math removed. Pangram~4 is scored on the text as written.}
\label{fig:fpr-tpr}
\label{fig:mathfrac}
\vspace{-6pt}
\end{figure}

In Figure~\ref{fig:fpr-tpr}, we additionally report the results of our detectors at FPRs going as low as $0.001\%$.
Pangram~4 reports a false-positive rate of $0.004\%$ in its technical report \citep{pangram4report}, and we accordingly mark the values of Pangram in the figure as well.
Even at these stricter operating points,
which are often needed for large-scale commercial deployment,
we find that \textsc{SnapJudge} still detects above $0.67$ of proofs and $0.80$ of prose passages,
far above the detection rate of Pangram.
In addition, \textsc{SnapJudge} still performs well at low resolution.
Appendix~\ref{app:resolution} shows that after the images are down-scaled threefold, \textsc{SnapJudge} still reaches a detection rate above $80\%$.
Overall these results establish that:
(i) AI-generated math-heavy writing, even from frontier models, can be detected with a high degree of accuracy;
and (ii) image-based detection of text is a viable alternative to pure text-based detection when
formatting may carry authorship signals.

\subsection{Detection across mathematical density and proof quality}\label{sec:mathfrac}
Despite the high accuracy of our detectors,
a natural question to ask is: what causes AI-generated math to be detectable?
Our hypothesis is that, just like AI-generated prose carries signals based on word and phrase choices,
mathematical writing also involves stylistic choices which do not impact the validity or
correctness of the final output.
In this section we rule out two alternative explanations.


First, we measure the amount of mathematics in a passage, which we define as the share of characters of the passage's LaTeX source that lie inside math mode.
Figure~\ref{fig:mathfrac} gives the detection rates across math-fraction bins, with proofs split at the quartiles of the test set and prose split at $5\%$ and $10\%$ (most prose passages have ${<}20\%$ math fraction).
We find that, while the detectors all degrade as the math fraction rises, they remain above a $75\%$ detection
rate across the entire range.
Moreover, removing the math content altogether or replacing it with a placeholder for the LaTeX representation,
leads to sharp drop in detection rate.
Together these results suggest that, in addition to text,
math equations and formulas also carry valuable signal which can help detect AI authorship.

Next, we examine whether our detectors rely on semantic signals which may be affected by the
quality of the models involved,
such as correctness or equivalence to human writing.
We sample 1,100 proof pairs and 1,000 prose pairs generated by the five frontier models and present each pair to GPT-5.6 Sol \citep{gpt56} 
to rate their similarity on a scale from $1$ to $10$. A pair comprises one human and one AI-generated proof  (or prose, depending on the condition). 
We classify pairs with scores of $1$--$5$ as non-equivalent and those with scores of $6$--$10$ as equivalent.
For proof pairs, the judge is additionally asked to judge whether each proof is correct or not.
We show in Appendix~\ref{app:judge-validation} that  GPT-5.6 Sol 
can distinguish correct proofs from incorrect ones (Table~\ref{tab:judge-validation}).

\begin{wrapfigure}{r}{0.50\textwidth}
\vspace{0pt}
\centering
\includegraphics[width=\linewidth]{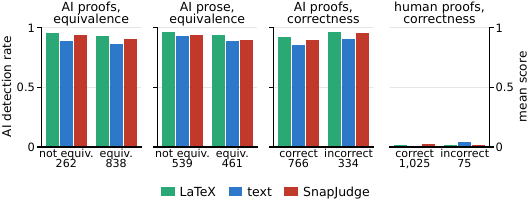}
\vspace{-18pt}
\caption{Detection rate at $0.1\%$ FPR on subsets defined by GPT-5.6 (left three panels) and mean score of human-written proofs (right). Numbers below bars count the passages.}
\label{fig:judge}
\vspace{-1.0\baselineskip}
\end{wrapfigure}
As Figure~\ref{fig:judge} shows, all three detectors still detect most AI-generated proofs that are judged to be correct and most AI passages that are judged to be equivalent to the original human-written proof, with rates only a few points below those on incorrect proofs or non-equivalent passages.
This small gap may arise because LLMs, when uncertain how to proceed, become less constrained by the content and fall back on characteristic stylistic patterns.
On the human side, errors alone do not make a proof look AI-generated: human-written proofs that are judged to be incorrect are not flagged as AI by any of the three detectors at $0.1\%$ FPR.
Taken together, these results show that the detectors respond to how a passage is written rather than to its correctness or content.



\subsection{Generalization to unseen generators and domains}
\label{sec:ood}

\begin{wraptable}{R}{0.34\textwidth}
\centering
\vspace{-10pt}
\footnotesize
\renewcommand{\arraystretch}{1.0}
\setlength{\tabcolsep}{2.5pt}
\begin{tabular}{@{}lcccc@{}}
\toprule
 & \multicolumn{2}{c}{\shortstack{all 15 models\\in training}} & \multicolumn{2}{c}{\shortstack{leave-one-\\model-out}} \\
\cmidrule(lr){2-3}\cmidrule(lr){4-5}
9B detector & Proofs & Prose & Proofs & Prose \\
\midrule
LaTeX & 0.855 & \textbf{0.889} & 0.745 & \textbf{0.820} \\
text & 0.754 & 0.849 & 0.669 & 0.808 \\
SnapJudge & \textbf{0.880} & 0.847 & \textbf{0.812} & 0.806 \\
\bottomrule
\end{tabular}
\caption{Generalization of the 9B detectors at $0.1\%$ FPR.}
\label{tab:generalization}

\vspace{-10pt}
\end{wraptable}
We hold out each of GPT-5.1, o4-mini, Claude Sonnet~5 and Gemini~3.1 Pro in turn, train a detector on the remaining fourteen models and test it on the held-out model, using the Qwen3.5-9B backbone to keep the training cost manageable.
Table~\ref{tab:generalization} shows that our detectors still perform well on these held-out models.
The LaTeX detector loses the most, $11.0$ points on proofs and $6.9$ on prose against $6.8$ and $4.1$ for \textsc{SnapJudge}.
We conjecture that the LaTeX detector partly relies on markup patterns that differ among models, such as the choice between \textbackslash( and \$ for inline mathematics (Section~\ref{sec:main}), and such patterns do not generalize to an unseen model.
Compilation renders these variants identically, so \textsc{SnapJudge} relies on what remains visible, such as the wording and the typeset formulas, which varies less across models and therefore, we speculate, it generalizes better.

%
\begin{wraptable}[9]{r}{0.42\textwidth}
\centering
\vspace{-18pt}
\scriptsize
\setlength{\tabcolsep}{2pt}
\renewcommand{\arraystretch}{0.9}
\setlength{\abovecaptionskip}{3pt}
\resizebox{\linewidth}{!}{%
\begin{tabular}{@{}lcccc@{}}
\toprule
TPR at 0.1\% FPR & MIRAGE & Peer review & RAID & EditLens \\
\midrule
Binoculars & 0.242 & 0.540 & 0.771 & 0.194 \\
Fast-DetectGPT & 0.033 & 0.459 & 0.479 & 0.332 \\
EditLens-Llama & 0.995 & 0.964 & \textbf{0.849} & \textbf{1.000} \\
EditLens-RoBERTa & 0.851 & 0.684 & 0.537 & 0.995 \\
Benchmark paper & 0.777$^{\ddagger}$ & 0.765 & 0.927$^{\ddagger}$ & 1.000 \\
\midrule
Ours (text) & \textbf{0.997} & \textbf{1.000} & 0.823 & 0.998 \\
SnapJudge & 0.995 & \textbf{1.000} & 0.842 & \textbf{1.000} \\
\bottomrule
\end{tabular}}
\caption{Cross-domain evaluation on four public benchmarks. $^{\ddagger}$: at $5\%$ FPR.}
\label{tab:xd-summary}
\vspace{-10pt}
\end{wraptable}
To test robustness against distribution shift---where it is conceivable that human-written text might be confidently misclassified  as AI \citep{rethinking2026}---we test our detectors on other public datasets to show that they perform equally well on unfamiliar domains.
Table~\ref{tab:xd-summary} reports the results on MIRAGE \citep{mirage}, the peer-review data of ICLR and NeurIPS \citep{intellabsbench}, the subset of RAID generated by six chat models \citep{raid}, and the AI-generated part of the EditLens dataset \citep{editlens}.
Under our own default thresholds, the false-positive rate is very low on every test set, no higher than $0.02\%$, and the
detection rates are generally on-par with the best methods from the respective papers.
\looseness=-1

We also check whether the scores of our detectors are well calibrated and find that \textsc{SnapJudge} performs best, with the lowest expected calibration error \citep{guo2017calibration} on both our own test sets and the four external benchmarks of Table~\ref{tab:xd-summary} (Appendix~\ref{app:reliability}).


\subsection{Detection of AI-edited passages}
\label{sec:mixed}
We test our detectors on human-written texts with varying degrees of AI editing, using three public general-domain datasets, EditLens \citep{editlens}, MIRAGE \citep{mirage} and APT-Eval \citep{apteval}, together with our own multi-level AI-editing test set (Section~\ref{sec:dataset}).

\begin{wraptable}{r}{0pt}
\vspace{0pt}
\centering
\centering
\footnotesize
\setlength{\tabcolsep}{3pt}
\begin{tabular}{@{}lc@{}}
\toprule
Detector & Spearman $\rho$ \\
\midrule
Binoculars & 0.420 \\
Fast-DetectGPT & 0.345 \\
EditLens-Llama & \textbf{0.651} \\
EditLens-RoBERTa & 0.561 \\
\midrule
Ours (text) & 0.624 \\
SnapJudge & 0.649 \\
\bottomrule
\end{tabular}

\captionof{table}{Spearman $\rho$ with the degree of AI editing.}
\label{tab:spearman}
\vspace{-\baselineskip}
\end{wraptable}
Table~\ref{tab:spearman} reports the Spearman correlation between the detector scores and the degree of AI editing, computed on each of the datasets above separately and then averaged.
We use the same procedure as EditLens \citep{editlens} for measuring editing degree, based on the cosine distance between the Linq-Embed-Mistral \citep{linqembed} embeddings of the human-written original and the edited passage.
A cosine distance below $0.03$ is recorded as $0$, one above $0.15$ as $1$, and values between $0.03$ and $0.15$ are interpolated linearly, which gives the editing degree.
The detection rate on each dataset and editing condition is listed in Table~\ref{tab:mixed} (Appendix~\ref{app:mixed}).

On these mixed datasets there is no fixed boundary between positives and negatives, and where to draw it depends on how much AI editing one is willing to accept. Although \textsc{SnapJudge} is trained only on binary labels, the Spearman correlation between its scores and the degree of AI editing is about as high as that of EditLens, which is trained to predict this degree, and far higher than those of the zero-shot detectors. A user can therefore set their threshold accordingly based on how much AI editing they are willing to accept.

\FloatBarrier
\Needspace{20\baselineskip}
\subsection{AI-generated text on arXiv  over time}
\label{sec:trend}
\begin{wrapfigure}{R}{0.45\textwidth}
\vspace{-\baselineskip}
\centering
\includegraphics[width=\linewidth]{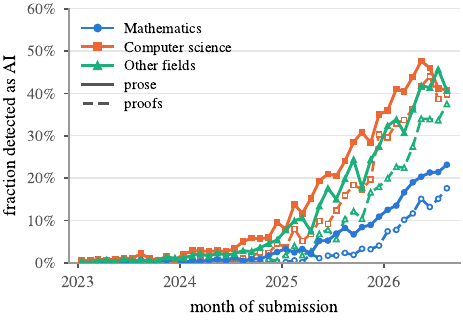}
\vspace{-15pt}
\setlength{\abovecaptionskip}{2pt}
\caption{Share of arXiv passages flagged as AI by the LaTeX detector (tuned to have a $0.01\%$ FPR).}
\label{fig:trend}
\vspace{-9pt}
\end{wrapfigure}
As a concrete application,
we use the LaTeX detector to study the share of AI-generated text in arXiv papers over time.
We sample arXiv papers with theorem proofs from every month between 2018 and 2026 and take one randomly sampled proof and prose passage from each, grouping the papers into mathematics, computer science and other fields (Appendix~\ref{app:trend-monthly}).
The detector uses the threshold for a false-positive rate of $0.01\%$ on the calibration pool (Section~\ref{sec:dataset}).
Detection rates on text before ChatGPT release confirms that FPR is extremely low in practice.
Among the proofs and prose passages of 2018, 2019 and 2021, only $0.008\%$ of the proofs and $0.003\%$ of the prose are falsely flagged.

Figure~\ref{fig:trend} shows that AI entered mathematics relatively late.
Its use in mathematics is nevertheless growing rapidly, and since the beginning of 2026 the curves for both prose and proofs have become markedly steeper. By August 2026, $17.6\%$ of proofs are detected as AI-generated.
AI-generated math-heavy text is therefore no longer a rare case, which underscores the need for detectors that work reliably on math-heavy content.

\FloatBarrier

\section{Conclusion}

In this work, we curated a dataset comprising coupled human-written and AI-generated proofs and prose. We explored training detectors using three different input representations: LaTeX, text extracted from PDF, and rendered images. Although it was unclear a priori if AI-generated math-heavy text would be detectable, our experiments consistently show that it is. In fact, our detectors are viable for practical use and outperform the commercial state-of-the-art detector, Pangram~4, on scientific text.
Further, we demonstrated that our detectors 
generalize to unseen models and domains, and its scores are correlated with the extent of AI editing.
Lastly, we analyzed recent trends related to AI-use in arXiv papers, and find AI use in mathematical content to be rapidly rising. Our work lays the foundation for interesting future work around building detectors that use multiple representations, and devising mechanisms to quantify AI contributions at a finer-level of detail. 



\section{Limitations}
\label{sec:limitations}
While this work shows that AI-written mathematics remains detectable and that our detectors perform very well at distinguishing purely human-written from purely AI-generated text, our work still has some limitations.
First, our detectors cannot yet reliably separate AI-polished text from fully AI-generated text.
On the human-written, but AI-polished (H-AI) subset of the peer-review data from \citet{saha2026policiespermittingllmuse}, even with the threshold set at an extremely low false-positive rate of $0.005\%$, we flag about $17\%$ to $42\%$ of the reviews as AI (Table~\ref{tab:peer-review} in Appendix~\ref{app:peer-review}).
We find that tightening the false-positive threshold further
leads to a large drop in detection rates for the purely AI-generated categories (Table~\ref{tab:peer-review-sweep}).
Hence, distinguishing AI-polished text from both human-written and AI-generated text remains a challenge,
and we leave it as a direction for future work.

In addition, although \textsc{SnapJudge} directly reads the rendered page images, it is so far trained only on text passages of proofs or prose and cannot yet make judgments on more multimodal content such as figures and tables.
Combining the judgment of figures and tables with that of text is another direction for future work.

\section{AI Use Statement}
The research questions and ideas of this paper were proposed by the authors, and some of the ideas emerged from discussions with AI.
The experiments were designed and proposed by the authors, and Claude Opus 5, Fable 5 and Opus 5.5 helped write the code, schedule the training jobs and compute the experimental results.
All results were manually reviewed and checked.
The search for and organization of related work were done with the help of Opus 5.
For the vast majority of the main text, the authors wrote the first draft, which was polished by Opus 5 and Opus 5.5 and then further revised by the authors.
Some passages were written in Chinese, translated into English by Opus 5 and Opus 5.5, and then revised.
All figures and tables in the paper were generated by Opus 5 and Opus 5.5 and then reviewed by the authors.
The first draft of the appendix was mainly generated by Opus 5 and Opus 5.5, and the authors reviewed and revised it.
All prompts used in the experiments were generated by Opus 5 and Fable 5 from key points provided by the authors, who also reviewed them.
References were imported into the bib file by the authors.
We take full responsibility for the final content of this work, including text, claims and artifacts produced with the aid of generative AI.

\section*{Acknowledgements}
This work was partially supported by a grant from the Cosmos Institute.
We thank Kevin Guo, Rounak Saha, Bruce Donald, Patrick Lopatto and members of the DukeNLP group for discussions and providing feedback on the detector.
The experiments with Pangram were carried out using a Pangram Research Award of \$5K credits to Bhuwan Dhingra, and we thank Pangram for their support.
The findings and conclusions presented here are of the authors alone and do not necessarily reflect the views of our sponsors.

\bibliography{iclr2027_conference}
\bibliographystyle{iclr2027_conference}

\clearpage
\appendix

\FloatBarrier

\section{Training Details and Hyperparameters}\label{app:hparams}

\begin{table}[H]
\centering
\footnotesize
\setlength{\tabcolsep}{3pt}
\renewcommand{\arraystretch}{1.15}
\begin{tabular}{@{}l*{5}{>{\raggedright\arraybackslash}p{0.145\linewidth}}@{}}
\toprule
27B models & LaTeX & Text & \textsc{SnapJudge} & OCR & AI-fraction estimator \\
\midrule
backbone & Qwen3.5-27B & Qwen3.5-27B & Qwen3.5-27B & Qwen3.5-27B & Qwen3.5-27B \\
input & LaTeX source, 512-token windows, stride 256 & PyMuPDF text, 512-token windows, stride 256 & rendered image, visual tokens of $32\times32$ pixels & OCR text of the rendered image, 512-token windows, stride 256 & rendered image \\
instruction & none & none & ``Determine whether this mathematical proof was written by a human or generated by AI.'' & none & ``Estimate how much of this scientific page excerpt was written by AI.'' \\
head output & 2 classes & 2 classes & 2 classes & 2 classes & 5 buckets \\
LoRA rank $r$ & 8 & 8 & 16 & 8 & 16 \\
learning rate & $10^{-4}$ & $10^{-4}$ & $2\times10^{-4}$ & $10^{-4}$ & $2\times10^{-4}$ \\
effective batch size & 16 & 16 & 64 & 16 & 64 \\
epochs & up to 3, best on
validation & up to 3, best on
validation & up to 3, best on
validation & up to 3, best on
validation & up to 3, best on
validation \\
seeds & 42, 7 & 42, 7 & 42, 7 & 42 & 42 \\
hardware, wall-clock & 4 RTX Pro 6000, 18 h & 4 H200, 12 h & 4 H200, 10 h & 4 H200, 5 h & 8 H200, 4.3 h \\
\bottomrule
\end{tabular}

\vspace{6pt}
\begin{tabular}{@{}l*{3}{>{\raggedright\arraybackslash}p{0.2\linewidth}}@{}}
\toprule
9B models & LaTeX & Text & Image \\
\midrule
backbone & Qwen3.5-9B & Qwen3.5-9B & Qwen3.5-9B \\
input & LaTeX source, 512-token windows, stride 256 & PyMuPDF text, 512-token windows, stride 256 & rendered image, visual tokens of $32\times32$ pixels \\
instruction & none & none & as \textsc{SnapJudge} 27B \\
head output & 2 classes & 2 classes & 2 classes \\
LoRA rank $r$ & 8 & 8 & 16 \\
learning rate & $10^{-4}$ & $10^{-4}$ & $2\times10^{-4}$ \\
effective batch size & 16 & 16 & 64 \\
epochs & up to 3, best on validation & up to 3, best on validation & up to 3, best on
validation \\
seeds & 7 (all 15 models), 42 and 7 (held out) & 7 (all 15 models), 42 and 7 (held out) & 7 \\
hardware, wall-clock & 2 H200, $\approx$5h & 2 H200, $\approx$5h & 2 H200, $\approx$5h \\
\bottomrule
\end{tabular}
\caption{Training settings of the 27B models behind our main results (top) and of the 9B detectors of Section~\ref{sec:ood} (bottom). }
\label{tab:hparams}
\end{table}
\FloatBarrier

\FloatBarrier
\section{Detection Rates by Generator Family}
\label{app:family}
Table~\ref{tab:unified-full} expands Table~\ref{tab:unified-main}: it lists, for every detector, the detection rate on each of the five frontier families that the main table averages into its last column.

\begin{table}[H]
\centering
\small
\renewcommand{\arraystretch}{1.0}
\setlength{\tabcolsep}{3pt}
\resizebox{0.81\linewidth}{!}{%
\begin{tabular}{@{}l@{\hskip 5pt}l@{\hskip 7pt}ccccc@{\hskip 8pt}ccccc@{}}
\toprule
 & & \multicolumn{5}{c}{Proofs} & \multicolumn{5}{c}{Prose} \\
\cmidrule(lr){3-7}\cmidrule(lr){8-12}
 & rep. & GPT-4.1 & GPT-5.1 & o4-mini & Claude & Gemini & GPT-4.1 & GPT-5.1 & o4-mini & Claude & Gemini \\
\midrule
\multicolumn{12}{c}{\textbf{LaTeX}} \\
\multicolumn{12}{c}{\emph{Shallow baselines (features + logistic regression)}} \\
\midrule
stylometric & LaTeX & 0.000 & 0.000 & 0.000 & 0.000 & 0.000 & 0.026 & 0.002 & 0.001 & 0.008 & 0.030 \\
char TF-IDF & LaTeX & 0.227 & 0.030 & 0.093 & 0.065 & 0.088 & 0.661 & 0.119 & 0.145 & 0.098 & 0.418 \\
word TF-IDF & LaTeX & \textbf{0.404} & \textbf{0.268} & \textbf{0.333} & \textbf{0.292} & \textbf{0.243} & \textbf{0.729} & \textbf{0.229} & \textbf{0.211} & \textbf{0.164} & \textbf{0.460} \\
\midrule
\multicolumn{12}{c}{\emph{Off-the-shelf detectors}} \\
\midrule
Binoculars & LaTeX & 0.014 & 0.000 & 0.000 & 0.000 & 0.014 & 0.004 & 0.000 & 0.000 & 0.000 & 0.006 \\
Fast-DetectGPT & LaTeX & 0.035 & 0.012 & 0.000 & 0.000 & 0.016 & 0.020 & 0.002 & 0.000 & 0.008 & 0.026 \\
EditLens-Llama & LaTeX & 0.004 & 0.004 & 0.000 & 0.004 & 0.024 & 0.814 & 0.119 & 0.105 & 0.194 & 0.471 \\
EditLens-RoBERTa & LaTeX & 0.039 & 0.004 & 0.011 & 0.019 & 0.039 & 0.285 & 0.035 & 0.019 & 0.041 & 0.199 \\
\addlinespace[2.5pt]
Pangram v3.3.2 & LaTeX & 0.098 & 0.043 & 0.389 & 0.095 & 0.045 & 0.860 & 0.546 & 0.716 & 0.333 & 0.450 \\
\quad\textit{calibrated} & LaTeX & 0.339 & 0.226 & 0.719 & 0.339 & 0.220 & 0.946 & 0.695 & 0.936 & 0.485 & 0.649 \\
\addlinespace[2.5pt]
Pangram 4 & LaTeX & 0.355 & 0.307 & 0.527 & 0.307 & 0.319 & 0.950 & 0.814 & 0.929 & 0.546 & 0.747 \\
\quad\textit{calibrated} & LaTeX & 0.876 & 0.890 & 0.957 & 0.898 & 0.756 & 0.990 & 0.938 & 0.990 & 0.773 & 0.848 \\
\midrule
\multicolumn{12}{c}{\emph{Ours, 9B}} \\
\midrule
leave-one-model-out & LaTeX & -- & 0.736 & 0.944 & 0.763 & 0.536 & -- & 0.908 & 0.976 & 0.639 & 0.755 \\
\addlinespace[2pt]
all 15 models & LaTeX & 0.920 & 0.907 & 0.959 & 0.851 & 0.702 & 0.990 & 0.936 & 0.985 & 0.792 & 0.845 \\
\addlinespace[2pt]
all 15 models, 0.01\% FPR & LaTeX & 0.790 & 0.730 & 0.842 & 0.691 & 0.545 & 0.982 & 0.884 & 0.962 & 0.702 & 0.744 \\
\midrule
\multicolumn{12}{c}{\emph{Ours, 27B}} \\
\midrule
all 15 models & LaTeX & 0.962 & 0.966 & 0.989 & 0.935 & 0.835 & 0.995 & 0.978 & 0.997 & 0.885 & 0.930 \\
\addlinespace[2pt]
all 15 models, 0.01\% FPR & LaTeX & 0.845 & 0.912 & 0.952 & 0.856 & 0.666 & 0.991 & 0.951 & 0.990 & 0.819 & 0.844 \\
\addlinespace[2pt]
all 15 models, 0.001\% FPR & LaTeX & 0.639 & 0.745 & 0.835 & 0.689 & 0.485 & 0.977 & 0.876 & 0.966 & 0.713 & 0.757 \\
\midrule
\multicolumn{12}{c}{\textbf{Text and image}} \\
\multicolumn{12}{c}{\emph{Shallow baselines (features + logistic regression)}} \\
\midrule
stylometric & text & 0.000 & 0.000 & 0.000 & 0.000 & 0.000 & 0.010 & 0.002 & 0.002 & 0.006 & 0.011 \\
char TF-IDF & text & 0.202 & 0.087 & 0.032 & 0.183 & 0.220 & \textbf{0.596} & 0.094 & 0.107 & 0.088 & \textbf{0.305} \\
word TF-IDF & text & \textbf{0.343} & \textbf{0.240} & \textbf{0.081} & \textbf{0.252} & \textbf{0.272} & 0.578 & \textbf{0.121} & \textbf{0.130} & \textbf{0.107} & 0.297 \\
\midrule
\multicolumn{12}{c}{\emph{Off-the-shelf detectors}} \\
\midrule
Binoculars & text & 0.000 & 0.000 & 0.000 & 0.000 & 0.005 & 0.002 & 0.002 & 0.001 & 0.002 & 0.000 \\
Fast-DetectGPT & text & 0.002 & 0.002 & 0.000 & 0.000 & 0.000 & 0.016 & 0.000 & 0.001 & 0.004 & 0.011 \\
EditLens-Llama & text & 0.000 & 0.000 & 0.000 & 0.002 & 0.000 & 0.611 & 0.077 & 0.079 & 0.120 & 0.341 \\
EditLens-RoBERTa & text & 0.009 & 0.002 & 0.005 & 0.017 & 0.007 & 0.217 & 0.028 & 0.024 & 0.034 & 0.106 \\
\addlinespace[2.5pt]
Pangram 4 & text & 0.278 & 0.347 & 0.387 & 0.351 & 0.259 & 0.870 & 0.763 & 0.745 & 0.443 & 0.646 \\
\quad\textit{calibrated} & text & 0.628 & 0.628 & 0.663 & 0.757 & 0.604 & 0.950 & 0.907 & 0.949 & 0.711 & 0.838 \\
\midrule
\multicolumn{12}{c}{\emph{Ours, 9B, text}} \\
\midrule
leave-one-model-out & text & -- & 0.775 & 0.658 & 0.679 & 0.564 & -- & 0.893 & 0.898 & 0.676 & 0.765 \\
\addlinespace[2pt]
all 15 models & text & 0.877 & 0.780 & 0.742 & 0.774 & 0.718 & 0.992 & 0.896 & 0.898 & 0.766 & 0.835 \\
\addlinespace[2pt]
all 15 models, 0.01\% FPR & text & 0.693 & 0.561 & 0.456 & 0.568 & 0.557 & 0.964 & 0.787 & 0.806 & 0.675 & 0.693 \\
\midrule
\multicolumn{12}{c}{\emph{SnapJudge, 9B}} \\
\midrule
leave-one-model-out & SnapJudge & -- & 0.944 & 0.775 & 0.859 & 0.668 & -- & 0.893 & 0.920 & 0.660 & 0.752 \\
\addlinespace[2pt]
all 15 models & SnapJudge & 0.933 & 0.943 & 0.866 & 0.926 & 0.784 & 0.980 & 0.893 & 0.950 & 0.726 & 0.818 \\
\addlinespace[2pt]
all 15 models, 0.01\% FPR & SnapJudge & 0.861 & 0.869 & 0.729 & 0.858 & 0.633 & 0.969 & 0.834 & 0.891 & 0.599 & 0.719 \\
\midrule
\multicolumn{12}{c}{\emph{Ours, 27B, text}} \\
\midrule
all 15 models & text & 0.913 & 0.918 & 0.848 & 0.928 & 0.780 & 0.982 & 0.938 & 0.953 & 0.832 & 0.895 \\
\addlinespace[2pt]
all 15 models, 0.01\% FPR & text & 0.827 & 0.825 & 0.752 & 0.866 & 0.665 & 0.969 & 0.906 & 0.923 & 0.757 & 0.835 \\
\addlinespace[2pt]
all 15 models, 0.001\% FPR & text & 0.662 & 0.659 & 0.601 & 0.748 & 0.507 & 0.959 & 0.852 & 0.859 & 0.680 & 0.725 \\
\midrule
\multicolumn{12}{c}{\emph{SnapJudge, 27B}} \\
\midrule
all 15 models & SnapJudge & 0.958 & 0.961 & 0.900 & 0.951 & 0.818 & 0.989 & 0.954 & 0.971 & 0.838 & 0.890 \\
\addlinespace[2pt]
all 15 models, 0.01\% FPR & SnapJudge & 0.886 & 0.886 & 0.764 & 0.880 & 0.690 & 0.982 & 0.914 & 0.936 & 0.755 & 0.825 \\
\addlinespace[2pt]
all 15 models, 0.001\% FPR & SnapJudge & 0.754 & 0.771 & 0.562 & 0.751 & 0.532 & 0.966 & 0.836 & 0.872 & 0.627 & 0.699 \\
\bottomrule
\end{tabular}}
\caption{Detection rate for each frontier family at $0.1\%$ FPR on
proofs and narrative prose; rows marked $0.01\%$ and $0.001\%$ FPR repeat our detectors at those FPRs. }
\label{tab:unified-full}
\end{table}

Table~\ref{tab:fpr-tiers} lists the values plotted in Figure~\ref{fig:fpr-tpr}.

\begin{table}[H]
\centering
\footnotesize
\renewcommand{\arraystretch}{1.0}
\setlength{\tabcolsep}{5pt}
\begin{tabular}{@{}llcccc@{}}
\toprule
 & & \multicolumn{2}{c}{Proofs} & \multicolumn{2}{c}{Prose} \\
\cmidrule(lr){3-4}\cmidrule(lr){5-6}
Detector & FPR & Frontier avg. & Open avg. & Frontier avg. & Open avg. \\
\midrule
LaTeX & 0.1\% & 0.938 & 0.920 & 0.957 & 0.943 \\
 & 0.05\% & 0.916 & 0.900 & 0.945 & 0.938 \\
 & 0.01\% & 0.846 & 0.849 & 0.919 & 0.924 \\
 & 0.005\% & 0.824 & 0.832 & 0.907 & 0.918 \\
 & 0.001\% & 0.678 & 0.745 & 0.858 & 0.893 \\
\midrule
text & 0.1\% & 0.877 & 0.889 & 0.920 & 0.921 \\
 & 0.05\% & 0.855 & 0.873 & 0.913 & 0.916 \\
 & 0.01\% & 0.787 & 0.828 & 0.878 & 0.901 \\
 & 0.005\% & 0.752 & 0.811 & 0.862 & 0.894 \\
 & 0.001\% & 0.636 & 0.747 & 0.815 & 0.871 \\
\midrule
SnapJudge & 0.1\% & 0.918 & 0.933 & 0.928 & 0.932 \\
 & 0.05\% & 0.888 & 0.918 & 0.915 & 0.926 \\
 & 0.01\% & 0.821 & 0.876 & 0.882 & 0.910 \\
 & 0.005\% & 0.782 & 0.852 & 0.856 & 0.903 \\
 & 0.001\% & 0.674 & 0.800 & 0.800 & 0.883 \\
\midrule
Pangram 4, LaTeX & 0.1\% (recalibrated) & 0.875 & -- & 0.908 & -- \\
 & 0.004\% (own labels) & 0.363 & -- & 0.797 & -- \\
\midrule
Pangram 4, text & 0.1\% (recalibrated) & 0.656 & -- & 0.871 & -- \\
 & 0.004\% (own labels) & 0.324 & -- & 0.694 & -- \\
\bottomrule
\end{tabular}
\caption{Values behind Figure~\ref{fig:fpr-tpr}: detection rate of the 27B detectors at five false-positive rates, as the frontier models average plotted in the figure and as the average over the ten open-source models, together with the Pangram~4 values on the text layer plotted in the figure and the corresponding values on the LaTeX source; $0.004\%$ is the false-positive rate that Pangram~4 reports for its own labels.}
\label{tab:fpr-tiers}
\end{table}

\FloatBarrier
\section{Detection Rates by Passage Length}
\begin{figure}[H]
\centering
\includegraphics[width=0.95\textwidth]{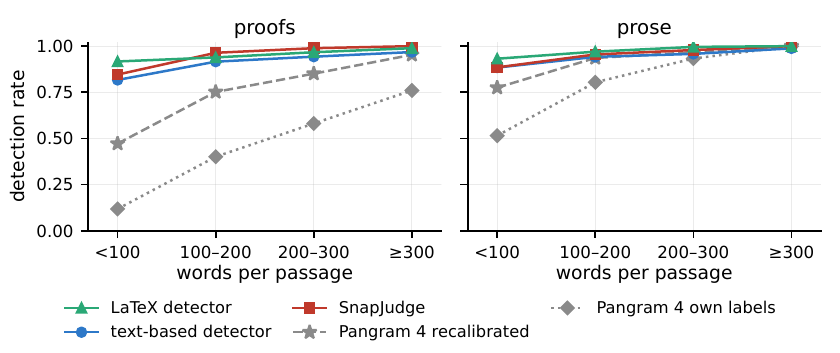}
\caption{Detection rate at $0.1\%$ FPR by passage length in words, frontier models average; Pangram~4 on the text input.}
\label{fig:length}
\end{figure}

Figure~\ref{fig:length} splits the frontier test passages of Table~\ref{tab:unified-main} by the length of the AI-generated passage, and keeps the thresholds of Table~\ref{tab:unified-main}.
Our detectors lose some accuracy on short passages but still detect more than $80\%$ of proofs and $88\%$ of prose below 100 words.
Pangram~4 depends much more on length.

\label{app:length}

\FloatBarrier
\section{Detection Rates on AI-Edited Datasets}
\label{app:mixed}
Table~\ref{tab:mixed} lists the detection rates of all detectors for every dataset and editing condition.

\begin{table}[H]
\centering
\normalsize
\renewcommand{\arraystretch}{1.05}
\setlength{\tabcolsep}{3pt}
\begin{tabular}{@{}l@{\hskip 6pt}cccccc@{}}
\toprule
 & \shortstack{EditLens\\Llama} & \shortstack{EditLens\\RoBERTa} & \shortstack{Fast-\\DetectGPT} & Binoculars & \shortstack{Ours,\\text} & SnapJudge \\
\midrule
\multicolumn{7}{c}{\emph{EditLens test set}} \\
\midrule
AI-edited & 0.593 & 0.382 & 0.025 & 0.009 & 0.204 & 0.249 \\
\midrule
\multicolumn{7}{c}{\emph{MIRAGE}} \\
\midrule
polish & 0.706 & 0.254 & 0.002 & 0.015 & 0.439 & 0.393 \\
rewrite & 0.757 & 0.248 & 0.002 & 0.013 & 0.398 & 0.354 \\
\midrule
\multicolumn{7}{c}{\emph{APT-Eval}} \\
\midrule
polish, extreme minor & 0.078 & 0.026 & 0.015 & 0.046 & 0.030 & 0.029 \\
10\% of sentences polished & 0.061 & 0.026 & 0.006 & 0.034 & 0.020 & 0.019 \\
polish, minor & 0.129 & 0.030 & 0.013 & 0.064 & 0.044 & 0.032 \\
50\% of sentences polished & 0.260 & 0.066 & 0.003 & 0.043 & 0.074 & 0.052 \\
polish, slight major & 0.456 & 0.104 & 0.011 & 0.091 & 0.162 & 0.142 \\
polish, major & 0.651 & 0.260 & 0.014 & 0.182 & 0.440 & 0.408 \\
\midrule
\multicolumn{7}{c}{\emph{Our multi-level AI-editing test set, proofs}} \\
\midrule
sentence-level rewrite & 0.011 & 0.020 & 0.000 & 0.002 & 0.005 & 0.007 \\
paragraph rewrite & 0.015 & 0.018 & 0.008 & 0.010 & 0.076 & 0.077 \\
two-stage generation & 0.192 & 0.222 & 0.232 & 0.145 & 0.891 & 0.937 \\
\midrule
\multicolumn{7}{c}{\emph{Our multi-level AI-editing test set, prose}} \\
\midrule
sentence-level rewrite & 0.051 & 0.015 & 0.004 & 0.001 & 0.036 & 0.017 \\
paragraph rewrite & 0.071 & 0.014 & 0.008 & 0.005 & 0.131 & 0.067 \\
two-stage  generation & 0.659 & 0.209 & 0.048 & 0.027 & 0.956 & 0.946 \\
\bottomrule
\end{tabular}
\caption{Detection rate on AI-edited passages. Off-the-shelf detectors are held at $0.1\%$ FPR; our detectors are at $0.01\%$ FPR. }
\label{tab:mixed}
\end{table}

\FloatBarrier

\section{Detection of AI-Polished Peer Reviews}
\label{app:peer-review}
We test our detectors on the peer-review benchmark of \citet{saha2026policiespermittingllmuse}, which contains human-written reviews (H), reviews written by an LLM from the paper and the reviewing guidelines (AI-BP), from the same inputs plus reviewing advice (AI-EP) or from key points extracted from a human review (AI-HI), and human-written reviews polished by an LLM that sees only the review (H-AI).
The easy subset is generated with GPT-4o and Llama-3.3-70B, and the hard subset with GPT-5, Gemini 2.5 Pro, Gemma-3-27B, Llama-3.1-70B and Qwen3-30B.
All three detectors read the plain text of each review, and \textsc{SnapJudge} reads it rendered on a single page.
The thresholds are those at a $0.005\%$ false-positive rate on our calibration pools.

We find that our detectors have no false positives on the human-written subset
and generally close to $100\%$ detection rate on the AI-$^\ast$ split of the test data (Table~\ref{tab:peer-review}).
The one category where our detector struggles is the H-AI split,
where human-written reviews are polished using an LLM,
for which the expected label is ``Human'' but are frequently labeled as ``AI''
by our detector.
The labeling of such passages is inherently subjective,
and dependent on the extent of polishing performed by the LLM;
\citet{saha2026policiespermittingllmuse} marked them as ``Human''
to conform with the policies adopted by several publishing venues which
restrict LLM use during reviewing to polishing only.
In Table~\ref{tab:peer-review-sweep}, we further experiment with
tightening the threshold for flagging reviews as AI:
we find that as the fraction of flagged H-AI reviews drops to $0$,
there is also a catastrophic drop in the detection rate of AI-$^\ast$
reviews,
suggesting that our detectors struggle at separating these categories.

\begin{table}[H]
\centering
\small
\renewcommand{\arraystretch}{1.05}
\setlength{\tabcolsep}{4pt}
\begin{tabular}{@{}l cccc cccc c@{}}
\toprule
 & \multicolumn{4}{c}{Easy} & \multicolumn{4}{c}{Hard} & \\
\cmidrule(lr){2-5}\cmidrule(lr){6-9}
 & AI-BP & AI-EP & AI-HI & H-AI & AI-BP & AI-EP & AI-HI & H-AI & H \\
\midrule
Pangram 3.0$^{\dagger}$ & 100.0 & 100.0 & 100.0 & 3.0 & 97.0 & 99.3 & 92.6 & 3.1 & 0.0 \\
GPTZero$^{\dagger}$ & 96.7 & 93.3 & 91.0 & 3.0 & 96.0 & 95.8 & 89.4 & 3.4 & 1.0 \\
\midrule
Ours, LaTeX & 100.0 & 100.0 & 99.9 & 44.3 & 99.9 & 100.0 & 99.95 & 35.6 & 0.0 \\
Ours, text & 100.0 & 100.0 & 100.0 & 45.9 & 100.0 & 100.0 & 99.95 & 40.4 & 0.0 \\
\textsc{SnapJudge} & 100.0 & 99.6 & 99.9 & 21.7 & 98.3 & 98.3 & 96.0 & 15.0 & 0.0 \\
\bottomrule
\end{tabular}
\caption{Percentage of reviews classified as AI on the test split of the peer-review benchmark of \citet{saha2026policiespermittingllmuse}. AI-BP, AI-EP and AI-HI are reviews written by an LLM, H-AI are human-written reviews polished by an LLM, and H are human-written reviews. Our detectors use the thresholds at a $0.005\%$ false-positive rate on our calibration pools. $^{\dagger}$Reported by \citet{saha2026policiespermittingllmuse},where a review counts as positive only when the entire review is labelled as AI-generated.}
\label{tab:peer-review}
\end{table}

\begin{table}[H]
\centering
\small
\renewcommand{\arraystretch}{1.05}
\setlength{\tabcolsep}{5pt}
\begin{tabular}{@{}l ccc ccc@{}}
\toprule
 & \multicolumn{3}{c}{Easy} & \multicolumn{3}{c}{Hard} \\
\cmidrule(lr){2-4}\cmidrule(lr){5-7}
H-AI flagged & LaTeX & Text & \textsc{SnapJudge} & LaTeX & Text & \textsc{SnapJudge} \\
\midrule
$\le 20\%$ & 98.6 & 99.0 & 100.0 & 97.1 & 98.7 & 98.0 \\
$\le 10\%$ & 92.9 & 90.2 & 98.8 & 67.2 & 65.5 & 87.6 \\
$\le 5\%$ & 74.5 & 50.3 & 92.0 & 29.9 & 16.9 & 66.1 \\
$\le 1\%$ & 8.1 & 0.8 & 56.6 & 5.1 & 1.9 & 12.0 \\
$0\%$ & 0.1 & 0.0 & 12.3 & 0.4 & 0.1 & 0.5 \\
\bottomrule
\end{tabular}
\caption{Detection rate of reviews written by an LLM (AI-BP, AI-EP and AI-HI pooled) on the test split of the benchmark of \citet{saha2026policiespermittingllmuse}, when the threshold of each detector is raised until at most the given share of the $2{,}802$ polished reviews (H-AI, easy and hard pooled) is flagged. }
\label{tab:peer-review-sweep}
\end{table}

\FloatBarrier
\section{Detection of Newer Models}
\label{app:newmodels}
While we were preparing this paper, two stronger models, GPT-6 and Claude Opus~5.5, were released, and we therefore also test our detectors on them.
For GPT-6 astra and Claude Opus~5.5 we generate 100 proof pairs and 100 prose pairs each from arXiv papers of 2021, following the procedure of Section~\ref{sec:dataset}.
To compare with the performance of Pangram~4, we use the thresholds at a $0.005\%$ false-positive rate on our calibration pools, which is close to the $0.004\%$ that Pangram~4 reports for its own labels \citep{pangram4report}.
Table~\ref{tab:newmodels} shows two main findings.
First, Pangram~4 performs very poorly on both models and labels at most a quarter of the passages as AI or mixed, and the detection rates of our detectors also drop clearly compared with other models (Table~\ref{tab:fpr-tiers}).
Second, retraining with only a small amount of data from these models already mitigates this difficulty.
Limited by our API budget, we add only 269 passages generated by GPT-6 astra and 881 generated by Claude Opus~5.5 to the training data of the LaTeX detector and slightly fewer to the other two detectors, which is only $1.5\%$ of the training data of each detector.
After retraining with these data, the detection rate already improves clearly.
At this threshold none of the human-written passages of these test sets is flagged, either before or after retraining.

\begin{table}[H]
\centering
\footnotesize
\renewcommand{\arraystretch}{1.0}
\setlength{\tabcolsep}{6pt}
\begin{tabular}{@{}lllccc@{}}
\toprule
Input & Passage & Model & Before & After & Pangram~4 \\
\midrule
LaTeX & Proofs & GPT-6 astra & 0.395 & 0.789 & 0.084 (0.011) \\
 & & Claude Opus~5.5 & 0.429 & 0.684 & 0.122 (0.000) \\
 & Prose & GPT-6 astra & 0.465 & 0.530 & 0.230 (0.120) \\
 & & Claude Opus~5.5 & 0.415 & 0.420 & 0.250 (0.120) \\
\midrule
Text & Proofs & GPT-6 astra & 0.274 & 0.645 & 0.108 (0.022) \\
 & & Claude Opus~5.5 & 0.391 & 0.615 & 0.146 (0.021) \\
 & Prose & GPT-6 astra & 0.405 & 0.547 & 0.211 (0.116) \\
 & & Claude Opus~5.5 & 0.422 & 0.469 & 0.219 (0.115) \\
\midrule
\textsc{SnapJudge} & Proofs & GPT-6 astra & 0.199 & 0.753 & -- \\
 & & Claude Opus~5.5 & 0.427 & 0.594 & -- \\
 & Prose & GPT-6 astra & 0.221 & 0.642 & -- \\
 & & Claude Opus~5.5 & 0.271 & 0.604 & -- \\
\bottomrule
\end{tabular}
\caption{Detection rate on the two newest models at a $0.005\%$ false-positive rate on our calibration pools. \emph{Before} is the detector of the main paper, which has seen no data from these models (mean of two seeds), and \emph{After} is the detector retrained with data from the new models added (one seed). For Pangram~4 we count a passage as detected when its own label is AI or mixed and give the share labelled AI in parentheses. Pangram~4 was run only on the AI-generated passages.}
\label{tab:newmodels}
\end{table}

\section{AI-Generated Image Detectors on Rendered Pages}
\label{app:pixel}

Existing image detectors ask whether an image was synthesized by an image generation model.
Our rendered pages are all real compiler output, and the two sides differ only in who wrote the text, so these detectors should carry no signal on our task.
Table~\ref{tab:pixel} confirms this. We score every page of both batteries with five open-source detectors, averaging over the crops of each document.
All of them sit at chance level.

\begin{table}[H]
\centering
\small
\begin{tabular}{@{}lcc@{}}
\toprule
Detector & Proof & Prose \\
\midrule
CNNDetection \citep{wang2020cnn} & 0.55 & 0.54 \\
UnivFD \citep{univfd}            & 0.49 & 0.54 \\
NPR \citep{npr}                  & 0.42 & 0.47 \\
SAFE \citep{safeimg}             & 0.53 & 0.48 \\
Community Forensics \citep{communityforensics} & 0.51 & 0.49 \\
\midrule
SnapJudge (ours)                 & 0.99 & 0.99 \\
\bottomrule
\end{tabular}
\caption{ AUROC of AI-generated image detectors and of our
detector on the rendered images.}
\label{tab:pixel}
\end{table}

\FloatBarrier

\section{Monthly Data of the arXiv Trend}
\label{app:trend-monthly}
For each month from 2018 to 2026 we sample arXiv papers that contain a theorem proof and take one proof and one prose passage from each paper.
We skip 2020, because most of our training data come from papers of that year, and use the months of 2018, 2019 and 2021 to check how often the detector flags text written before the release of ChatGPT.
Papers are assigned to mathematics, computer science (including eess and stat.ML) or other fields by their primary arXiv category.
Up to 2022, each month contains about 1,000 papers, most of them from mathematics.
From 2023 on, we add papers from computer science and other fields, so that each field contributes at least about 300 papers per month, and the months of 2026 contain more papers from every field.
Table~\ref{tab:trend-monthly} lists, for each month from 2023 to 2026, the number of papers and the shares of prose and proofs detected as AI in each field, and Table~\ref{tab:trend-field} summarizes all months by year.
\begin{table}[H]
\centering
\small
\setlength{\tabcolsep}{4.5pt}
\renewcommand{\arraystretch}{0.97}
\begin{tabular}{@{}lrrr@{\hskip 12pt}rrr@{\hskip 12pt}rrr@{}}
\toprule
 & \multicolumn{3}{c}{Mathematics} & \multicolumn{3}{c}{Computer science} & \multicolumn{3}{c}{Other fields} \\
\cmidrule(lr){2-4}\cmidrule(lr){5-7}\cmidrule(l){8-10}
month & $n$ & prose & proofs & $n$ & prose & proofs & $n$ & prose & proofs \\
\midrule
2023-01 & 626 & 0.0 & 0.2 & 600 & 0.5 & 0.0 & 322 & 0.3 & 0.0 \\
2023-02 & 583 & 0.0 & 0.2 & 600 & 0.5 & 0.0 & 304 & 0.0 & 0.0 \\
2023-03 & 638 & 0.0 & 0.0 & 600 & 0.8 & 0.0 & 381 & 0.5 & 0.0 \\
2023-04 & 645 & 0.2 & 0.2 & 600 & 0.7 & 0.2 & 299 & 0.7 & 0.3 \\
2023-05 & 566 & 0.2 & 0.2 & 600 & 0.8 & 0.0 & 367 & 0.3 & 0.0 \\
2023-06 & 613 & 0.2 & 0.0 & 600 & 1.2 & 0.2 & 349 & 0.9 & 0.0 \\
2023-07 & 611 & 0.0 & 0.0 & 600 & 1.0 & 0.2 & 370 & 0.5 & 0.0 \\
2023-08 & 692 & 0.7 & 0.0 & 600 & 2.2 & 0.2 & 308 & 0.6 & 0.0 \\
2023-09 & 653 & 0.3 & 0.2 & 600 & 1.0 & 0.3 & 353 & 0.6 & 0.0 \\
2023-10 & 616 & 0.3 & 0.0 & 600 & 0.7 & 0.2 & 397 & 0.5 & 0.0 \\
2023-11 & 652 & 0.6 & 0.0 & 600 & 1.5 & 0.0 & 373 & 1.3 & 0.0 \\
2023-12 & 648 & 0.5 & 0.0 & 600 & 0.7 & 0.5 & 366 & 1.1 & 0.0 \\
2024-01 & 676 & 0.0 & 0.0 & 600 & 2.0 & 0.5 & 353 & 1.1 & 0.0 \\
2024-02 & 583 & 0.5 & 0.0 & 600 & 2.8 & 0.0 & 405 & 1.7 & 0.0 \\
2024-03 & 642 & 0.3 & 0.2 & 600 & 2.8 & 0.5 & 378 & 2.1 & 0.3 \\
2024-04 & 674 & 0.6 & 0.0 & 600 & 2.7 & 1.0 & 381 & 1.6 & 0.3 \\
2024-05 & 594 & 0.8 & 0.0 & 600 & 2.8 & 0.5 & 432 & 2.3 & 0.0 \\
2024-06 & 595 & 0.5 & 0.0 & 600 & 2.7 & 0.3 & 382 & 1.8 & 0.0 \\
2024-07 & 622 & 1.3 & 0.0 & 600 & 3.3 & 0.8 & 460 & 2.2 & 0.2 \\
2024-08 & 654 & 0.5 & 0.2 & 600 & 5.0 & 1.0 & 387 & 2.8 & 0.3 \\
2024-09 & 662 & 0.9 & 0.2 & 600 & 5.8 & 1.5 & 374 & 2.7 & 0.0 \\
2024-10 & 550 & 0.9 & 0.0 & 600 & 5.7 & 2.5 & 560 & 3.6 & 0.2 \\
2024-11 & 636 & 1.7 & 0.6 & 600 & 6.0 & 2.3 & 461 & 4.6 & 0.9 \\
2024-12 & 666 & 2.6 & 0.3 & 600 & 9.5 & 4.5 & 409 & 5.4 & 0.5 \\
2025-01 & 639 & 3.3 & 0.2 & 600 & 8.0 & 3.7 & 388 & 7.7 & 1.8 \\
2025-02 & 531 & 2.4 & 0.6 & 600 & 13.8 & 8.0 & 323 & 9.9 & 4.0 \\
2025-03 & 583 & 3.3 & 0.2 & 600 & 11.7 & 5.2 & 388 & 10.6 & 1.8 \\
2025-04 & 628 & 2.2 & 2.1 & 600 & 15.2 & 6.8 & 336 & 7.4 & 2.7 \\
2025-05 & 637 & 5.2 & 1.1 & 600 & 19.3 & 9.8 & 351 & 13.4 & 6.8 \\
2025-06 & 534 & 5.2 & 1.7 & 600 & 21.0 & 9.2 & 358 & 17.6 & 7.8 \\
2025-07 & 637 & 6.9 & 1.7 & 600 & 20.5 & 12.3 & 426 & 15.0 & 5.6 \\
2025-08 & 644 & 8.2 & 2.3 & 600 & 24.0 & 15.8 & 348 & 19.8 & 10.3 \\
2025-09 & 643 & 6.7 & 1.9 & 600 & 28.5 & 18.3 & 435 & 24.4 & 12.2 \\
2025-10 & 546 & 8.4 & 3.3 & 600 & 30.8 & 17.3 & 547 & 17.7 & 10.4 \\
2025-11 & 593 & 8.9 & 3.2 & 600 & 28.3 & 19.7 & 420 & 24.3 & 16.7 \\
2025-12 & 615 & 10.9 & 4.1 & 600 & 35.0 & 30.3 & 484 & 27.5 & 18.0 \\
2026-01 & 2{,}462 & 12.5 & 7.4 & 1{,}091 & 36.0 & 29.6 & 516 & 32.4 & 20.0 \\
2026-02 & 2{,}264 & 13.5 & 7.8 & 1{,}741 & 41.0 & 32.9 & 520 & 33.8 & 22.7 \\
2026-03 & 2{,}200 & 16.6 & 10.1 & 1{,}184 & 40.5 & 33.7 & 534 & 30.7 & 22.5 \\
2026-04 & 2{,}404 & 19.0 & 11.6 & 1{,}343 & 43.9 & 36.3 & 568 & 36.3 & 27.5 \\
2026-05 & 2{,}293 & 20.3 & 15.1 & 1{,}816 & 47.6 & 41.7 & 579 & 41.8 & 34.0 \\
2026-06 & 2{,}444 & 21.3 & 13.2 & 1{,}302 & 46.0 & 44.0 & 563 & 41.4 & 33.9 \\
2026-07 & 2{,}913 & 21.4 & 15.1 & 1{,}291 & 41.2 & 38.7 & 600 & 45.7 & 33.7 \\
2026-08 & 3{,}119 & 23.1 & 17.6 & 1{,}303 & 40.8 & 39.8 & 600 & 40.7 & 37.5 \\
\bottomrule
\end{tabular}
\caption{Monthly data behind Figure~\ref{fig:trend}. For each field, $n$ is the number of papers, and the prose and proof columns give the share (\%) of passages detected as AI by the LaTeX detector at $0.01\%$ FPR.}
\label{tab:trend-monthly}
\end{table}

\begin{table}[H]
\centering
\small
\setlength{\tabcolsep}{6pt}
\begin{tabular}{@{}lrrrrrr@{}}
\toprule
 & \multicolumn{3}{c}{Prose} & \multicolumn{3}{c}{Proofs} \\
\cmidrule(lr){2-4}\cmidrule(l){5-7}
year & math & CS & other & math & CS & other \\
\midrule
2018 & 0.0 & 0.0 & 0.0 & 0.0 & 0.0 & 0.0 \\
2019 & 0.0 & 0.0 & 0.0 & 0.0 & 0.0 & 0.0 \\
2021 & 0.0 & 0.0 & 0.0 & 0.0 & 0.0 & 0.0 \\
2022 & 0.0 & 0.1 & 0.1 & 0.0 & 0.0 & 0.0 \\
2023 & 0.3 & 1.0 & 0.6 & 0.1 & 0.1 & 0.0 \\
2024 & 0.9 & 4.3 & 2.7 & 0.1 & 1.3 & 0.2 \\
2025 & 6.0 & 21.3 & 16.8 & 1.9 & 13.0 & 8.6 \\
2026 & 18.8 & 42.5 & 38.1 & 12.5 & 37.3 & 29.3 \\
\bottomrule
\end{tabular}
\caption{Share (\%) of passages detected as AI by the LaTeX detector at $0.01\%$ FPR, by year and field.}
\label{tab:trend-field}
\end{table}

\FloatBarrier

\section{SnapJudge at Lower Resolution}
\label{app:resolution}

\begin{figure}[H]
\centering
\includegraphics[width=0.55\linewidth]{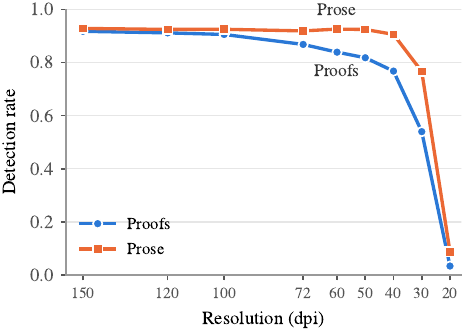}
\caption{Detection rate of \textsc{SnapJudge} on frontier-model test passages rendered at decreasing resolution.}
\label{fig:resolution}
\end{figure}

To test how \textsc{SnapJudge} copes with low-quality screenshots, we render the frontier test passages of Table~\ref{tab:unified-main} at resolutions from 150 dpi, the resolution used in training, down to 20 dpi, and score them with the threshold of Table~\ref{tab:unified-main} unchanged.
Figure~\ref{fig:resolution} shows that detection on prose barely changes down to 40 dpi, while detection on proofs declines gradually, from $91.8\%$ at 150 dpi to $81.8\%$ at 50 dpi.
Proofs are presumably more sensitive because subscripts and small symbols in the formulas become illegible first.
Below 40 dpi, detection drops sharply for both registers, as the text itself becomes hard to read.

\FloatBarrier

\section{Proof-Carrying Submissions at ICLR}
\label{app:iclr}
The statistics quoted in the introduction come from the public OpenReview records of the seven ICLR editions from 2020 to 2026.
A submission counts as proof-carrying if its PDF contains at least one passage that opens with a proof marker (``Proof.'' or ``Proof of \ldots'') and is closed by an end-of-proof mark.
Table~\ref{tab:iclr-years} gives, per year, the mean review score and the acceptance rate of such submissions against the other submissions of the same year.
Proof-carrying submissions receive higher review scores in every year, with all seven differences significant ($z$ from $3.8$ to $16.7$), and higher acceptance rates in every year, significant in six of the seven years (2021: $z=1.8$).

Moreover, the share of papers that carry proofs keeps rising, especially among theory-oriented papers (Table~\ref{tab:iclr-areas}).
The acceptance premium of carrying proofs is likewise concentrated in the nine areas, at $11$ to $15$ percentage points in every year ($p<2\times10^{-13}$).
\begin{table}[H]
\centering
\footnotesize
\renewcommand{\arraystretch}{1.05}
\setlength{\tabcolsep}{5pt}
\begin{tabular}{@{}lrccrccr@{}}
\toprule
 & & \multicolumn{3}{c}{mean review score} & \multicolumn{3}{c}{acceptance rate (\%)} \\
\cmidrule(lr){3-5}\cmidrule(lr){6-8}
Year & submissions & with proofs & without & $\Delta$ ($z$) & with proofs & without & $\Delta$ ($z$) \\
\midrule
2020 & 2,213 & 4.64 & 4.34 & $+0.30$ (3.8) & 36.5 & 29.1 & $+7.4$ (3.3) \\
2021 & 2,594 & 5.65 & 5.46 & $+0.19$ (4.4) & 35.9 & 32.1 & $+3.8$ (1.8) \\
2022 & 2,617 & 5.77 & 5.41 & $+0.36$ (7.0) & 47.2 & 39.3 & $+7.9$ (3.8) \\
2023 & 3,792 & 5.68 & 5.42 & $+0.26$ (5.4) & 46.7 & 40.1 & $+6.6$ (3.4) \\
2024 & 7,404 & 5.44 & 5.05 & $+0.39$ (11.8) & 46.0 & 36.2 & $+9.8$ (6.7) \\
2025 & 11,672 & 5.45 & 5.06 & $+0.39$ (14.8) & 47.2 & 40.4 & $+6.8$ (5.6) \\
2026 & 19,814 & 4.46 & 4.13 & $+0.33$ (16.7) & 39.6 & 34.2 & $+5.4$ (6.0) \\
\bottomrule
\end{tabular}
\caption{Review scores and acceptance rates of proof-carrying versus other ICLR submissions, 2020--2026.  $\Delta$ is the difference between proof-carrying and other submissions, with the two-sample $z$ statistic in parentheses; $|z|>1.96$ corresponds to significance at the $5\%$ level (two-sided).}
\label{tab:iclr-years}
\end{table}

\begin{table}[H]
\centering
\footnotesize
\renewcommand{\arraystretch}{1.05}
\setlength{\tabcolsep}{5pt}
\begin{tabular}{@{}>{\raggedright\arraybackslash}p{0.60\linewidth}ccc@{}}
\toprule
 & \multicolumn{3}{c}{share of submissions with proofs} \\
\cmidrule(lr){2-4}
Primary area & 2024 & 2025 & 2026 \\
\midrule
\multicolumn{4}{@{}l}{\emph{The nine high-proof areas (share $\ge 25\%$ in 2026)}} \\
\quad learning theory & 69.0\% & 71.5\% & 78.4\% \\
\quad causal reasoning & 49.5\% & 56.3\% & 58.5\% \\
\quad optimization & 50.6\% & 57.8\% & 57.2\% \\
\quad probabilistic methods (Bayesian methods, sampling, UQ) & 40.6\% & 42.4\% & 51.6\% \\
\quad learning on graphs and other geometries & 45.6\% & 45.9\% & 44.0\% \\
\quad reinforcement learning & 36.6\% & 41.6\% & 39.9\% \\
\quad other topics in machine learning & -- & 29.0\% & 29.8\% \\
\quad transfer, meta and lifelong learning & 18.4\% & 17.4\% & 25.6\% \\
\quad unsupervised, self-supervised, semi-supervised and supervised representation learning & 23.6\% & 25.2\% & 25.5\% \\
\quad \emph{the nine areas together} & 33.1\% & 39.9\% & 41.5\% \\
\midrule
All other areas & 13.3\% & 13.6\% & 14.2\% \\
\midrule
\multicolumn{4}{@{}l}{\emph{Acceptance gap of carrying proofs (percentage points)}} \\
\quad the nine areas together & $+15.4$ & $+13.0$ & $+11.1$ \\
\quad all other areas & $+4.2$ & $+4.4$ & $+4.9$ \\
\bottomrule
\end{tabular}
\caption{Proof-carrying submissions by research area, 2024--2026. }
\label{tab:iclr-areas}
\end{table}

\FloatBarrier

\section{Generator Models and Generation Prompts}\label{app:prompts}

\paragraph{Generator models.} The ten open-source generators are Qwen3-32B and Qwen3-235B \citep{qwen3}, Qwen2.5-72B \citep{qwen25}, Llama-3.3-70B \citep{llama3}, Llama-4-Scout \citep{llama4}, DeepSeek-R1-Distill-Qwen-32B and DeepSeek-R1-Distill-Llama-70B \citep{deepseekr1}, GLM-4.5-Air \citep{glm45}, Phi-4 \citep{phi4} and gpt-oss-120B \citep{gptoss}, all run locally with vLLM \citep{vllm}.
The five frontier generators are GPT-4.1 \citep{gpt41}, GPT-5.1 \citep{gpt51}, o4-mini \citep{o4mini}, Claude Sonnet 5 \citep{claudesonnet5} and Gemini 3.1 Pro \citep{gemini31}, accessed through their APIs.
Each target passage is generated by exactly one model.

\paragraph{Generation cost.} Table~\ref{tab:gencost} lists the number of passages from each frontier API that are used in the training, validation and test splits, and the corresponding cost. Every request carries the full LaTeX source of the paper as context, so the cost is dominated by input tokens and comes to about \$0.03 per passage; the whole frontier half of the dataset cost about \$390. The ten open-source generators ran on our own GPUs and incurred no API cost.

\paragraph{Judging cost.} The LLM judge of Section~\ref{sec:mathfrac} (GPT-5.6 Sol \citep{gpt56}, reasoning effort set to high) read 1,100 proof pairs and 1,000 prose pairs, about 3.3M input tokens and 0.7M output tokens including reasoning tokens, for about \$40.

\begin{table}[H]
\centering
\footnotesize
\begin{tabular}{@{}lrrr@{}}
\toprule
Generator & Proof passages & Prose passages & Cost (USD) \\
\midrule
GPT-4.1 & 1,477 & 1,244 & $\approx$90 \\
GPT-5.1 & 1,487 & 1,600 & $\approx$85 \\
o4-mini & 801 & 1,600 & $\approx$60 \\
Claude Sonnet 5 & 1,053 & 1,147 & $\approx$70 \\
Gemini 3.1 Pro & 1,040 & 1,045 & $\approx$95 \\
\midrule
Total & 5,858 & 6,636 & $\approx$390 \\
\midrule
GPT-5.6 Sol (judge, Sec.~\ref{sec:mathfrac}) & 1,100 pairs & 1,000 pairs & $\approx$40 \\
\bottomrule
\end{tabular}
\caption{API cost of the frontier-model passages used in the dataset, and of the LLM judge.}
\label{tab:gencost}
\end{table}

\paragraph{System prompt for proofs.}
\begin{quote}\small\ttfamily\raggedright
You are an expert mathematician completing the LaTeX source of a research paper.
You will be given the paper's title, abstract, macro definitions, and the full LaTeX source of the paper with one proof removed and its position marked; the theorem statement remains in place immediately before the marker.
Your task is to write the missing proof.

Rules:\\ 1.
Write a complete, rigorous proof of the stated theorem.\\ 2.
Write the proof so that it reads as a seamless continuation of the paper's own source.
Follow the excerpt's writing style, notation, phrasing, and LaTeX conventions as closely as you can: use the same inline/display math delimiters, the same environment names, and the paper's macros where their meaning is clear from the excerpt.\\ 3.
Output ONLY the proof body in raw LaTeX.
Do NOT include \textbackslash{}begin\{proof\} or \textbackslash{}end\{proof\}, do not restate the theorem, and do not add headings, greetings, or commentary of any kind.\\ 4.
Aim for approximately $N$ words.
\end{quote}

\paragraph{System prompt for prose.}
\begin{quote}\small\ttfamily\raggedright
You are an expert academic writer completing the LaTeX source of a research paper.
ONE paragraph of running text has been removed from the body and replaced with a marker.
You are given the paper's title, abstract, macro definitions, optionally its references, and the FULL LaTeX source with that one paragraph missing.
Write the missing paragraph.

Rules:\\ 1.
Write a single paragraph that fits seamlessly at the marker: it must follow from the text before it and lead into the text after it.\\ 2.
Match the paper's own writing style, notation, terminology, phrasing, and LaTeX conventions: the same inline math delimiters, the same custom macros, the same citation commands (cite using keys that appear elsewhere in the paper), and the same register.\\ 3.
Output ONLY the missing paragraph as raw LaTeX running text.
Do NOT add a section heading, list, theorem/proof environment, figure, table, or display-math block; do not restate the surrounding text; and do not add any greeting, label, or commentary.\\ 4.
Aim for approximately $N$ words.
\end{quote}

\paragraph{Prompts for the multi-level AI-editing test set.} 

\begin{quote}\small\ttfamily\raggedright
\textrm{\emph{Sentence-level rewrite.}}\\ Rewrite this sentence from an academic paper to improve fluency and clarity.
Keep the meaning of the sentence unchanged.
Output only the rewritten sentence in Latex, nothing else.\\ Sentence: \{s\}
\end{quote}

\begin{quote}\small\ttfamily\raggedright
\textrm{\emph{Paragraph rewrite.}}\\ Rewrite the following passage from a mathematical paper to improve its fluency and clarity.
Keep the meaning and all mathematical content unchanged.
Output only the rewritten passage in LaTeX, nothing else.\\ Passage: \{s\}
\end{quote}

\begin{quote}\small\ttfamily\raggedright
\textrm{\emph{Two-stage free generation, stage 1 (outline).}}\\ List every logical step and key point of the following passage as a concise bullet outline, in your own words.
You may use mathematical symbols.
Do not copy phrases from the original text.
Output only the outline.\\ Passage: \{s\}
\end{quote}

\begin{quote}\small\ttfamily\raggedright
\textrm{\emph{Two-stage free generation, stage 2 (writing; proofs).}}\\ Write a complete, rigorous mathematical proof in standard LaTeX that follows this outline.
Output only the proof body, no preamble.\\ Outline: \{s\}
\end{quote}

\begin{quote}\small\ttfamily\raggedright
\textrm{\emph{Two-stage free generation, stage 2 (writing; prose).}}\\ Write a well-structured paragraph of an academic mathematics paper in LaTeX that covers exactly the following points.
Output only the paragraph.\\ Points: \{s\}
\end{quote}

\FloatBarrier
\section{Prompts of the Equivalence Judge}\label{app:equiv-prompt}

\paragraph{System prompt for proofs.}
\begin{quote}\small\ttfamily\raggedright
You are an expert research mathematician acting as a referee.
You will be given a theorem statement taken from a mathematics paper and two proofs of it, labelled Proof A and Proof B.
The material is an excerpt: definitions, notation and results cited from elsewhere in the paper are not shown, and you should assume that such cited facts exist and are as the authors say.

Your task has two parts.

1.
Validity.
For each proof, judge whether it is a correct and complete proof of the statement as written.
Answer "yes" if the argument is correct and no essential step is missing, "partial" if the strategy is sound but a genuine gap or error remains, and "no" if the argument does not establish the statement.
Do not penalise brevity, informality, omitted routine verifications or reliance on cited results.

2.
Similarity of mathematical content.
Compare what the two proofs actually do, not how they are written.
Ignore wording, sentence order, verbosity, the names of variables and the LaTeX formatting.
Focus on the overall strategy, the key steps, the intermediate claims and the results invoked.
Rate the similarity on a scale from 1 to 10.
Be generous: two proofs that reach the conclusion by the same overall route deserve a high score even if they differ in details, in the amount of detail shown, or in the order of the steps.
Reserve low scores for proofs whose mathematical strategies genuinely differ.
Use the following anchors:\\ 10 = the same proof, the same argument with the same key steps, differing only in presentation;\\ 8-9 = the same argument with small differences of content, such as an extra or omitted routine step, a different order of the same steps, a different but equivalent computation, or a different choice of which routine facts to spell out;\\ 6-7 = the same overall strategy and mostly the same key steps, but one key step is carried out differently or a different intermediate result is invoked;\\ 4-5 = the strategies overlap only in part, such as sharing the main idea but reaching it through substantially different intermediate steps;\\ 2-3 = different strategies or different key steps, although both arguments are plausible;\\ 1 = fundamentally different arguments, or one of the proofs is not a proof of the statement at all.

Do not try to guess who wrote either proof and do not let style, polish or length influence the similarity rating.
Reply with one JSON object and nothing else, in a fenced \textasciigrave\textasciigrave\textasciigrave json block, with exactly these fields:\\ \{"A\_valid": "yes\textbar partial\textbar no", "B\_valid": "yes\textbar partial\textbar no", "similarity": \textless integer 1-10\textgreater, "same\_strategy": \textless true\textbar false\textgreater, "key\_difference": "\textless one sentence naming the most important difference in content, or 'none' if the proofs are essentially the same\textgreater"\}
\end{quote}

\paragraph{User message for proofs.}
\begin{quote}\small\ttfamily\raggedright
THEOREM (\{thm\_kind\}):\\ \{statement\}

PROOF A:\\ \{proof\_a\}

PROOF B:\\ \{proof\_b\}
\end{quote}

\paragraph{System prompt for prose.}
\begin{quote}\small\ttfamily\raggedright
You are an expert academic reviewer.
You will be given two versions of one paragraph from a mathematics paper, labelled Paragraph A and Paragraph B.
Both are meant to occupy the same position in the paper, and the material is an excerpt: the rest of the paper, its definitions and its cited results are not shown, and you should assume that such cited facts exist and are as the authors say.

Your task is to judge how similar the two paragraphs are in content.
Compare what the paragraphs say, not how they say it.
Ignore wording, sentence order, verbosity, the choice of words, citation commands and LaTeX formatting.
Focus on the claims made, the facts and results stated, the objects and quantities discussed, and the arguments or motivations given.
Rate the similarity on a scale from 1 to 10.
Be generous: two paragraphs that make the same points deserve a high score even if they differ in detail, in emphasis, in the amount of detail given, or in the order in which the points are made.
Reserve low scores for paragraphs whose content genuinely differs.
Use the following anchors:\\ 10 = the same content, the same points made in the same way, differing only in presentation;\\ 8-9 = the same points with small differences of content, such as an extra or omitted minor detail, a different order of the same points, or a different but equivalent way of stating a fact;\\ 6-7 = mostly the same points, but one substantive point is added, dropped or stated differently;\\ 4-5 = the paragraphs overlap only in part, such as sharing the main point but otherwise discussing different things;\\ 2-3 = largely different content, although both paragraphs are plausible at this position in the paper;\\ 1 = essentially unrelated content, or one of the paragraphs does not fit the position at all.

Do not try to guess who wrote either paragraph and do not let style, polish or length influence the rating.
Reply with one JSON object and nothing else, in a fenced \textasciigrave\textasciigrave\textasciigrave json block, with exactly these fields:\\ \{"similarity": \textless{}integer 1-10\textgreater{}, "key\_difference": "\textless{}one sentence naming the most important difference in content, or 'none' if the paragraphs are essentially the same\textgreater{}"\}
\end{quote}

\paragraph{User message for prose.}
\begin{quote}\small\ttfamily\raggedright
PARAGRAPH A:\\ \{para\_a\}

PARAGRAPH B:\\ \{para\_b\}
\end{quote}

\FloatBarrier
\section{Validation of the Proof Judge}\label{app:judge-validation}

\begin{table}[H]
\centering
\small
\begin{tabular}{@{}lcc@{}}
\toprule
 & Flawed proofs (17) & Control proofs (13) \\
\midrule
Mean severity & 7.08 & 4.23 \\
Standard deviation & 2.46 & 2.20 \\
Median & 8.0 & 4.0 \\
Severity $\geq 5$ & 12/17 (71\%) & 2/13 (15\%) \\
\bottomrule
\end{tabular}
\caption{Severity scores from 0 to 10 that GPT-5.6 Sol assigns to proofs from withdrawn ICLR 2026 submissions whose reviewers pointed out a flaw, and to control proofs that reviewers described as correct, averaged over the judge's samples for each proof. For this check the judge uses the scoring prompt given below. A flawed proof receives a higher score than a control proof with probability $0.81$ (one-sided Mann-Whitney test, $p=0.0016$).}
\label{tab:judge-validation}
\end{table}

\paragraph{System prompt.}
\begin{quote}\small\ttfamily\raggedright
You are verifying a single mathematical result taken from a machine learning paper. You will be shown the problem setting, the statements of results the proof relies on, and one statement together with its proof.

Your task is NOT to decide whether the proof "has a problem." It does. They all do. Every published proof contains steps that are compressed, hypotheses left implicit, and constants left unnamed. Asking whether a defect exists is a question whose answer is always yes, and it carries no information. What we need from you is a measurement of CONSEQUENCE: how much work would it take to make this proof correct?

WHAT YOU MUST DO, IN THIS ORDER

Step 1 -- Re-derive. Work through the proof line by line. Check each inference rather than reading for plausibility. Where the algebra is checkable numerically, check it numerically. For every inequality, limit, expectation, or asymptotic claim, ask what happens at the boundary of the stated parameter range.

Step 2 -- Enumerate, including trivia. List every imperfection you can find, from notational slips upward. Finding things is free here and proves nothing; we want the list so that we can see you looked. Three findings is a reasonable number. Do not stop at the first one and do not pad the list with restatements of the same issue.

Step 3 -- Rank, then score ONLY the worst. Severity does not accumulate. Three cosmetic slips are still cosmetic. The count of findings is never the score. If your top three findings all look serious, you have almost certainly split one issue into three; merge them.

Step 4 -- Attempt the repair. This is the step that determines your score, and it is the reason this task is not a matter of opinion.

THE PRIMARY SCORE: repair

\hspace*{1.0em}none\_needed\quad -- You searched and found only cosmetic or notational issues. A referee would not raise them.

\hspace*{1.0em}one\_line\quad -- There is a real gap, and you can close it. You MUST actually write the missing step or the missing hypothesis in the repair\_attempt field. If you cannot write it out, this is not the right level.

\hspace*{1.0em}nontrivial\quad -- You tried the one-line repair and it does not work. You MUST state in repair\_attempt what new argument is required AND why the short route fails. "It needs more detail" is not an answer.

\hspace*{1.0em}unfixable\quad -- The statement itself is false, or the proof's central construction cannot exist. You MUST exhibit or sketch a counterexample to the STATEMENT in repair\_attempt -- concrete values, or a family of instances, and what goes wrong. A counterexample to one STEP is not enough; a step can be replaced. If you cannot produce a counterexample to the statement, you are at nontrivial, not here.

Each level costs you something we can check. Do not claim a level whose price you have not paid.

THE SECONDARY SCORE: severity, 0 to 10\\
\hspace*{1.0em}0\quad no defect found after genuine effort\\
\hspace*{1.0em}1-2\quad cosmetic, notational, or pedantic; a referee would not mention it\\
\hspace*{1.0em}3-4\quad a real gap that a careful reader repairs in one line while reading\\
\hspace*{1.0em}5-6\quad a genuine hole needing a nontrivial new argument; the claim is still likely true\\
\hspace*{1.0em}7-8\quad the proof as written does not establish the claim, and the repair route is unclear\\
\hspace*{1.0em}9-10\quad the claim as stated is false, and you can exhibit or sketch a counterexample

Scoring 1 to 4 is the normal, expected, respectable outcome for a competently written proof. Do not inflate. But do not reflexively assign a low score either -- a low score must be earned by a genuine search that turned up only minor things, and your enumeration in step 2 must show that search.

WHAT YOU ARE NOT ASKED FOR\\
Do not report how confident you feel. We have measured self-reported confidence on this task and it carries no information; it will be discarded. Ground your answer in the mathematics and in the repair you did or could not do.

ON MATERIAL YOU WERE NOT GIVEN\\
You are shown an excerpt, not the whole paper. If a quantity looks undefined or a hypothesis looks missing, consider first that it is probably established elsewhere in the paper. Treat that as the default. Only count it as a defect when the proof's own logic requires something that could not plausibly be supplied elsewhere -- for example when the step is false under any reading, or when the proof itself asserts the missing fact rather than citing it. Do not build an objection on a hostile choice of parameters that the paper's setting would rule out.

EVIDENCE RULES\\
Every claimed defect must be anchored in text you COPY VERBATIM from the proof you were shown. Copy the substring exactly, including LaTeX. Do not paraphrase, do not reconstruct, do not normalise the notation. If you cannot copy a real substring, you do not have a defect to report.

OUTPUT\\
Return one JSON object and nothing else, inside a fenced code block marked json. Escape backslashes correctly so the object parses; LaTeX in your quotes must survive.

\{\\
\hspace*{1.0em}"findings": [\\
\hspace*{2.0em}\{"quote": "\textless{}verbatim substring from the proof\textgreater{}", "what\_is\_wrong": "\textless{}one sentence\textgreater{}"\}\\
\hspace*{1.0em}],\\
\hspace*{1.0em}"worst\_finding\_index": \textless{}integer index into findings, or -1 if none\textgreater{},\\
\hspace*{1.0em}"repair": "none\_needed" \textbar{} "one\_line" \textbar{} "nontrivial" \textbar{} "unfixable",\\
\hspace*{1.0em}"repair\_attempt": "\textless{}the missing step written out; or why the short route fails and what is needed; or the counterexample. Required unless repair is none\_needed.\textgreater{}",\\
\hspace*{1.0em}"severity": \textless{}integer 0-10\textgreater{},\\
\hspace*{1.0em}"conclusion\_status": "true\_as\_stated" \textbar{} "true\_after\_restatement" \textbar{} "probably\_true\_unproven" \textbar{} "false",\\
\hspace*{1.0em}"reasoning": "\textless{}why the worst finding sits at that repair level, in a few sentences\textgreater{}"\\
\}
\end{quote}

\paragraph{User message.}
\begin{quote}\small\ttfamily\raggedright
Problem setting.\\ \{setup\}

Statements this proof relies on.\\ - \{dependency 1\}\\ - \{dependency 2\}\\ \ldots

Statement and proof to evaluate.\\ \{statement and proof\}
\end{quote}

\FloatBarrier

\section{Calibration of the Detector Scores}
\label{app:reliability}
\begin{figure}[H]
\centering
\includegraphics[width=0.45\linewidth]{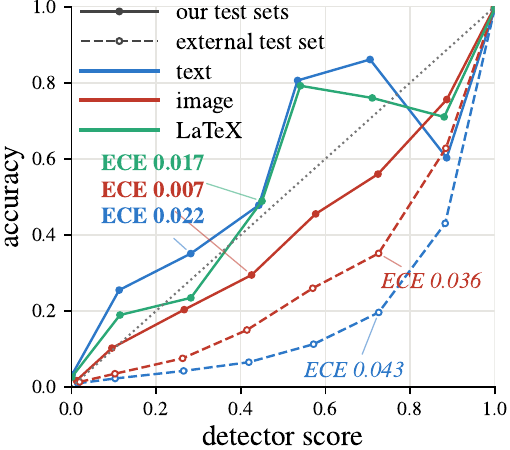}
\vspace{-3pt}
\caption{Reliability of the 27B detectors.}
\label{fig:reliability}
\end{figure}

Figure~\ref{fig:reliability} shows the reliability diagrams of the three 27B detectors.
We divide the scores reported by a detector into eight ranges (the detailed values are given in Table~\ref{tab:reliability}).
The horizontal coordinate of each point is the mean score of the texts in that range and the vertical coordinate is the share of AI samples among them.
Within every dataset the AI samples are first weighted to count as much as the human-written samples, so as to avoid the effect of unequal sample sizes.
The figure shows that SnapJudge is less miscalibrated than the text-based detector on both the in-domain and the external test sets, and that it alleviates the problem of the model being over-confident on unfamiliar human-written text.

We use the expected calibration error (ECE) \citep{guo2017calibration} to express the degree of miscalibration of a detector.
It is computed as
\begin{equation}
\mathrm{ECE} = \sum_{b=1}^{8} \frac{n_b}{N}\,\lvert y_b - x_b \rvert,
\label{eq:ece}
\end{equation}
where $x_b$ and $y_b$ are the horizontal and vertical coordinates of range $b$, $n_b$ is the weighted number of texts in it and $N$ is the total number of texts.

\vspace{-10pt}
\vspace{-10pt}
\begin{table}[H]
\centering
\small
\setlength{\tabcolsep}{7pt}
\renewcommand{\arraystretch}{0.85}
\begin{tabular}{@{}l cc cc cc@{}}
\toprule
 & \multicolumn{2}{c}{Text} & \multicolumn{2}{c}{SnapJudge} & \multicolumn{2}{c}{LaTeX} \\
\cmidrule(lr){2-3}\cmidrule(lr){4-5}\cmidrule(lr){6-7}
Score range & score & AI share & score & AI share & score & AI share \\
\midrule
\multicolumn{7}{@{}l}{\emph{Our test sets}} \\
0--0.05 & 0.001 & 0.032 & 0.012 & 0.017 & 0.001 & 0.026 \\
0.05--0.2 & 0.113 & 0.255 & 0.096 & 0.102 & 0.115 & 0.189 \\
0.2--0.35 & 0.282 & 0.351 & 0.267 & 0.203 & 0.282 & 0.234 \\
0.35--0.5 & 0.443 & 0.478 & 0.426 & 0.295 & 0.450 & 0.489 \\
0.5--0.65 & 0.534 & 0.806 & 0.577 & 0.455 & 0.541 & 0.792 \\
0.65--0.8 & 0.706 & 0.861 & 0.724 & 0.560 & 0.711 & 0.760 \\
0.8--0.95 & 0.887 & 0.603 & 0.887 & 0.756 & 0.881 & 0.710 \\
0.95--1 & 1.000 & 0.992 & 0.999 & 0.998 & 1.000 & 0.995 \\
ECE & \multicolumn{2}{c}{0.022} & \multicolumn{2}{c}{0.007} & \multicolumn{2}{c}{0.017} \\
\midrule
\multicolumn{7}{@{}l}{\emph{External test set}} \\
0--0.05 & 0.015 & 0.010 & 0.019 & 0.013 & -- & -- \\
0.05--0.2 & 0.103 & 0.022 & 0.103 & 0.035 & -- & -- \\
0.2--0.35 & 0.265 & 0.042 & 0.263 & 0.075 & -- & -- \\
0.35--0.5 & 0.419 & 0.065 & 0.415 & 0.150 & -- & -- \\
0.5--0.65 & 0.572 & 0.113 & 0.570 & 0.260 & -- & -- \\
0.65--0.8 & 0.726 & 0.196 & 0.726 & 0.351 & -- & -- \\
0.8--0.95 & 0.883 & 0.430 & 0.885 & 0.628 & -- & -- \\
0.95--1 & 0.999 & 0.991 & 0.999 & 0.995 & -- & -- \\
ECE & \multicolumn{2}{c}{0.043} & \multicolumn{2}{c}{0.036} & \multicolumn{2}{c}{--} \\
\bottomrule
\end{tabular}
\caption{Values behind Figure~\ref{fig:reliability}.}
\label{tab:reliability}
\end{table}

\section{AI-Fraction Estimator}\label{app:fraction-confusion}

To make \textsc{SnapJudge} more practical, we separately train a Qwen3.5-27B estimator to predict the fraction of the text in an image, measured by height, that is AI-generated, using the same fine-tuning procedure as for the detectors and a cross-entropy loss over five buckets. Its bucket accuracy on the test set is $0.885$.
The fraction estimator is trained on the real-page windows of Section~\ref{sec:dataset},  grouped by their AI fraction $p$ into the five buckets $(0,0.25)$, $[0.25,0.5)$, $[0.5,0.75)$, $[0.75,1)$ and $\{1\}$.
It outputs bucket probabilities $p_1,\dots,p_5$, the reported bucket is the most probable one, and the fraction estimate is $\hat\rho=\sum_j p_j m_j$ with $m=(0.125,0.375,0.625,0.875,1)$. It is only ever applied to passages that SnapJudge has already judged AI-generated.

Table~\ref{tab:fraction-confusion} evaluates it on 5{,}000 framed windows from the 2021 test papers, 1{,}000 per true bucket.
Part (a) shows the estimator alone.
Part (b) shows the deployed cascade, in which SnapJudge first decides at its $0.1\%$ false-positive threshold whether a window contains AI text, and only the windows it passes on are assigned a bucket.
Besides the bucket decision, we measure the error of the fraction estimate $\hat\rho$ as the mean absolute error $|\hat\rho-\rho|$ against the AI fraction $\rho$ of the window, reported per true bucket in the last column of each part.
The estimator alone is off by $0.055$ on average and the cascade by $0.051$ on the windows it passes.

\begin{table}[H]
\centering
\footnotesize
\setlength{\tabcolsep}{3pt}
\begin{minipage}[t]{0.45\linewidth}
\centering
\textbf{(a)} estimator alone, all AI-containing windows\\[3pt]
\begin{tabular}{@{}lrrrrr@{\hskip 6pt}r@{}}
\toprule
true $\backslash$ pred. & 1 & 2 & 3 & 4 & 5 & MAE \\
\midrule
1 $(0,.25)$ & \textbf{91.4} & 6.5 & 1.1 & 0.2 & 0.8 & 0.077 \\
2 $[.25,.5)$ & 2.8 & \textbf{91.6} & 4.8 & 0.2 & 0.6 & 0.058 \\
3 $[.5,.75)$ & 0.5 & 5.1 & \textbf{89.0} & 3.8 & 1.6 & 0.062 \\
4 $[.75,1)$ & 0.5 & 0.5 & 4.8 & \textbf{72.4} & 21.8 & 0.055 \\
5 (AI) & 0.1 & 0.2 & 1.0 & 0.6 & \textbf{98.1} & 0.023 \\
\midrule
all & & & & & & 0.055 \\
\bottomrule
\end{tabular}
\end{minipage}\hfill
\begin{minipage}[t]{0.53\linewidth}
\centering
\textbf{(b)} cascade: image-detector gate at 0.1\% FPR, then the estimator\\[3pt]
\begin{tabular}{@{}lrrrrrr@{\hskip 6pt}r@{}}
\toprule
true $\backslash$ pred. & passed & 1 & 2 & 3 & 4 & 5 & MAE \\
\midrule
1 $(0,.25)$ & 66.1 & \textbf{93.4} & 5.3 & 0.9 & 0.0 & 0.4 & 0.072 \\
2 $[.25,.5)$ & 81.2 & 2.2 & \textbf{93.1} & 4.2 & 0.1 & 0.4 & 0.057 \\
3 $[.5,.75)$ & 86.9 & 0.1 & 5.2 & \textbf{90.3} & 3.6 & 0.7 & 0.059 \\
4 $[.75,1)$ & 88.5 & 0.2 & 0.5 & 4.8 & \textbf{73.8} & 20.7 & 0.053 \\
5 (AI) & 88.4 & 0.1 & 0.2 & 0.9 & 0.6 & \textbf{98.2} & 0.021 \\
\midrule
all & 82.2 & & & & & & 0.051 \\
\bottomrule
\end{tabular}
\end{minipage}
\caption{The AI-fraction estimator on 5{,}000 framed test windows.}
\label{tab:fraction-confusion}
\end{table}

\begin{figure}[H]
\centering
\includegraphics[width=\linewidth]{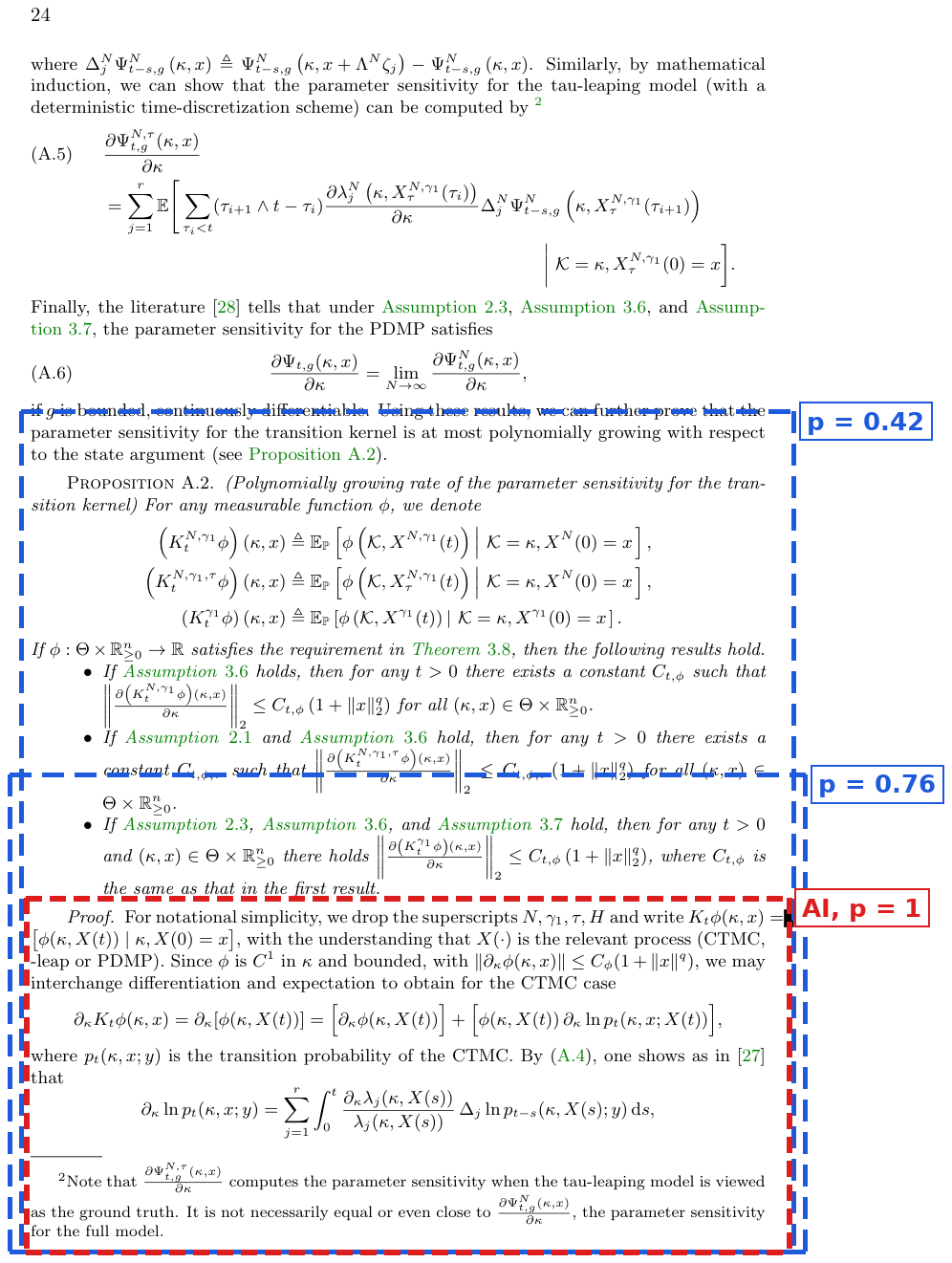}
\caption{A recompiled page of a 2021 test paper in which an AI-generated proof (dashed red box, $p=1$) is surrounded by the human-written text of the paper. The dashed blue boxes are two windows opened by the sampler on this page, each fully covering the AI passage together with part of the human-written context; their AI fraction $p$ is the height of the AI text divided by the height of all text inside the window.}
\label{fig:fraction-page}
\end{figure}

\FloatBarrier

\end{document}